\documentclass{article}

    \PassOptionsToPackage{numbers, compress}{natbib}
\usepackage[preprint]{neurips_data_2023}
\usepackage{url}            % simple URL typesetting
\usepackage{booktabs}       % professional-quality tables
\usepackage{amsfonts}       % blackboard math symbols
\usepackage{amsmath}
\usepackage{xcolor} 
\usepackage{hyperref}
\usepackage{graphicx,calc} 
\usepackage{makecell}
\usepackage{wrapfig}

\newcommand{\Rmnum}[1]{\expandafter\@slowromancap\romannumeral #1@}
\usepackage{bbm}
\usepackage{amssymb}
\usepackage{multirow}
\usepackage{fontawesome5}
\usepackage{nicematrix} 
\usepackage{pifont}
\newlength\myheight
\newlength\mydepth
\settototalheight\myheight{Xygp}
\definecolor{uclablue}{rgb}{0.15, 0.45, 0.68}
\definecolor{lightcoral}{rgb}{0.94, 0.5, 0.5}
\definecolor{lightgreen}{rgb}{0.56, 0.93, 0.56}
\definecolor{harvestgold}{rgb}{0.85, 0.57, 0.0}
\definecolor{brightlavender}{rgb}{0.75, 0.58, 0.89}
\definecolor{capri}{rgb}{0, 0.61, 0.94}
\definecolor{carminepink}{rgb}{0.92, 0.3, 0.26}
\definecolor{celadon}{rgb}{0.67, 0.88, 0.69}
\definecolor{darkpastelgreen}{rgb}{0.01, 0.75, 0.24}
\definecolor{cplus}{rgb}{1,0.42,0}
\definecolor{cminus}{rgb}{0,0.48,0.76}
\usepackage{cleveref}
\newcommand\refsec[1]{\hyperref[sec:#1]{§\ref{sec:#1}:~\textsc{#1}}}
\newcommand\refsecs[2]{\hyperref[sec:#1]{§\ref{sec:#1}:~\textsc{#1}}, \hyperref[sec:#2]{§\ref{sec:#2}:~\textsc{#2}}}

\providecommand{\keywords}[1]
{
  \small
  \textbf{\textit{Keywords---}} #1
}

\hypersetup{
    breaklinks,
    citecolor=uclablue,
    colorlinks=true,
    linkcolor=uclablue
}
\newcommand{\NA}{--}
\newcommand{\eg}{\emph{e.g.}}

\newcommand{\wikirecent}{WikiData$_{recent}$}
\newcommand{\wikicf}{WikiData$_{counterfact}$}
\newcommand{\wikibio}{WikiBio}
\newcommand{\zsre}{ZsRE}
\newcommand{\convsent}{Convsent}
\newcommand{\sani}{Sanitation}

\newcommand*\colourcheck[1]{%
  \expandafter\newcommand\csname #1check\endcsname{\textcolor{#1}{\ding{52}}}%
}
\newcommand*\colourcross[1]{%
  \expandafter\newcommand\csname #1cross\endcsname{\textcolor{#1}{\ding{55}}}%
}
\newcommand{\bcircle}{\textcolor{capri}{\faIcon[regular]{circle}}}
\newcommand{\bplus}{\textcolor{cplus}{\faIcon[regular]{plus}}}
\newcommand{\bminus}{\textcolor{cminus}{\faIcon[regular]{minus}}}
\colourcheck{blue}
\colourcheck{green}
\colourcheck{darkpastelgreen}
\colourcross{blue}
\colourcross{applegreen}
\colourcross{red}

\title{A Comprehensive Study of Knowledge Editing for Large Language Models}

\author{%
Ningyu Zhang\thanks{Equal Contribution.}~~, Yunzhi Yao$^{*}$, Bozhong Tian$^{*}$, Peng Wang$^{*}$, Shumin Deng$^{*}$, Mengru Wang, 
\\ 
\textbf{Zekun Xi, Shengyu Mao, Jintian Zhang, Yuansheng Ni, Siyuan Cheng, Ziwen Xu,} 
\\ 
\textbf{Xin Xu, Jia-Chen Gu, Yong Jiang, Pengjun Xie, Fei Huang, Lei Liang,} 
\\
\textbf{Zhiqiang Zhang, Xiaowei Zhu, Jun Zhou, Huajun Chen\thanks{Corresponding Author.}}
\\ 
\\
{ \vspace{0.2cm} Zhejiang University,~ National University of Singapore,~ \vspace{-0.2cm} }
\\
{University of California, Los Angeles,~ Ant Group,~ Alibaba Group} 
\\
\texttt{\{zhangningyu,yyztodd\}@zju.edu.cn}
\\
Project: \url{https://zjunlp.github.io/project/KnowEdit}
}

\begin{document}

\maketitle

% \vspace{2mm}
\setcounter{footnote}{0}

\begin{abstract}
% \vspace{-0.1cm}

Large Language Models (LLMs) have shown extraordinary capabilities in understanding and generating text that closely mirrors human communication. However, a primary limitation lies in the significant computational demands during training, arising from their extensive parameterization. This challenge is further intensified by the dynamic nature of the world, necessitating frequent updates to LLMs to correct outdated information or integrate new knowledge, thereby ensuring their continued relevance. Note that many applications demand continual model adjustments post-training to address deficiencies or undesirable behaviors. There is an increasing interest in efficient, lightweight methods for on-the-fly model modifications. To this end, recent years have seen a burgeoning in the techniques of \textbf{knowledge editing}  for LLMs, which aim to \emph{efficiently} modify LLMs' behaviors within specific domains while preserving overall performance across various inputs. In this paper, we first define the knowledge editing problem and then provide a comprehensive review of cutting-edge approaches. Drawing inspiration from educational and cognitive research theories \cite{Bruner1964TheCO,Bruner1960ThePO,jayashri2018knowledge}, we propose a unified categorization criterion that classifies knowledge editing methods into three groups: \textit{resorting to external knowledge}, \textit{merging knowledge into the model}, and \textit{editing intrinsic knowledge}. Furthermore, we introduce a new benchmark, \textbf{KnowEdit}, for a comprehensive empirical evaluation of representative knowledge editing approaches. Additionally, we provide an in-depth analysis of knowledge location, which can give a deeper understanding of the knowledge structures inherent within LLMs. Initially conceived as a means to steer LLMs efficiently, we hope that insights gained from knowledge editing research could shed light on the underlying knowledge mechanisms of LLMs. To facilitate future research, we have released an open-source framework, EasyEdit\footnote{\url{https://github.com/zjunlp/EasyEdit}. \\ \\  $\quad\quad$The contributions of the authors are detailed in \hyperref[sec:contributions]{\textsc{$\S$Contributions}}.}, which will enable practitioners to efficiently and flexibly implement knowledge editing for LLMs. Finally, we discuss several potential applications of knowledge editing, outlining its broad and impactful implications.

\end{abstract}
\vspace{-0.3cm}
\hspace{30pt}
{\keywords{natural language processing, large language models, knowledge editing}}

\newpage
{
  \hypersetup{linktoc=page}
  \tableofcontents
}
\newpage

\section{Introduction}
\label{sec:introduction}
Knowledge is a fundamental component of human intelligence and civilization \cite{DBLP:journals/aim/DavisSS93}. 
Its systematic structure empowers us to represent tangible entities or delineate principles through symbolic means, offering the capability to facilitate the articulation of intricate behaviors or tasks \cite{DBLP:conf/wsdm/Choi22,DBLP:journals/ai/ZhangLPKOFS22,manning2022human}.
Throughout our lives, we humans continuously gather an extensive wealth of knowledge and learn to adaptively apply it in various contexts.
The enduring exploration of the nature of knowledge and the processes by which we acquire, retain, and interpret it, continues to captivate scientists, which is not just a technical pursuit but a journey towards mirroring the nuanced complexities of human cognition, communication and intelligence \cite{DBLP:books/daglib/0067119,DBLP:conf/naacl/Liu0BPS19,DBLP:journals/cacm/HanZL21,jarrahi2023artificial,DBLP:journals/corr/abs-2312-02706}.

Recently, Large Language Models (LLMs) like GPT-4 \cite{DBLP:journals/corr/abs-2303-08774} have showcased a remarkable ability in Natural Language Processing (NLP) to retain a vast amount of knowledge, arguably surpassing human capacity \cite{DBLP:journals/corr/abs-2303-18223,sawicki2023state,DBLP:journals/corr/abs-2302-13971,DBLP:journals/corr/abs-2307-10169,hadi2023large,DBLP:journals/corr/abs-2303-11717,DBLP:journals/corr/abs-2304-13712,DBLP:journals/corr/abs-2306-13549,DBLP:journals/corr/abs-2308-11432,DBLP:journals/corr/abs-2309-07864,DBLP:journals/corr/abs-2304-08354,DBLP:journals/corr/abs-2303-17580,DBLP:journals/corr/abs-2307-04964,wang2023survey,DBLP:journals/corr/abs-2309-15402,DBLP:conf/emnlp/QinZ0CYY23,DBLP:journals/corr/abs-2311-11797,DBLP:journals/tmlr/WeiTBRZBYBZMCHVLDF22}. 
This achievement can be attributed to the way LLMs process and compress huge amounts of data \cite{lotfi2023non,liu2023grokking,DBLP:journals/corr/abs-2309-10668,DBLP:journals/corr/abs-2312-01700}, potentially forming more concise, coherent, and interpretable models of the underlying generative processes, essentially creating a kind of ``world model'' \cite{gurnee2023language,DBLP:journals/corr/abs-2311-05876,DBLP:journals/corr/abs-2306-12672}.
For example, \citet{dai-etal-2022-knowledge} have introduced the Knowledge Neuron (KN) thesis, which proposes that language models function similarly to key-value memories. 
Here, the multi-layer perceptron (MLP) weights in the core region \cite{DBLP:journals/corr/abs-2310-14928} may play a crucial role in recalling facts from the training corpus, suggesting a more structured and retrievable form of knowledge storage within LLMs \cite{geva-etal-2021-transformer,geva-etal-2022-transformer}.
Further insights come from the ability of LLMs to understand and manipulate complex strategic environments, whereas \citet{DBLP:conf/iclr/0002HBVPW23} has demonstrated that transformers trained for next-token prediction in board games such as Othello develop explicit representations of the game's state.   
\citet{DBLP:conf/iclr/PatelP22} have revealed that LLMs can track boolean states of subjects within given contexts and learn representations that reflect perceptual, symbolic concepts \cite{gurnee2023language,DBLP:journals/corr/abs-2309-14316,DBLP:journals/corr/abs-2309-14402,DBLP:conf/cvpr/YangPZJCY23}.
This dual capability indicates that LLMs can serve as extensive knowledge bases \cite{petroni-etal-2019-language,DBLP:journals/corr/abs-2008-09036,DBLP:conf/acl/WangL020,DBLP:conf/naacl/ZhongFC21,DBLP:conf/acl/CaoLHSYLXX20,DBLP:conf/emnlp/ZhaoZXZ022,DBLP:journals/tacl/DhingraCEGEC22,DBLP:journals/corr/abs-2204-06031,DBLP:journals/corr/abs-2303-07616,DBLP:conf/emnlp/YoussefKLSS23,DBLP:journals/corr/abs-2306-08302,DBLP:journals/tgdk/PanRKSCDJO0LBMB23}, not only storing vast amounts of information but also structuring it in ways that may mirror human cognitive processes.
 
However, LLMs have limitations like factual fallacy, potential generation of harmful content, and outdated knowledge due to their training cut-off \cite{DBLP:conf/emnlp/ZhangFCNW23,DBLP:journals/corr/abs-2311-05656,DBLP:conf/emnlp/YinH023,DBLP:journals/corr/abs-2310-05189}. 
Retraining to correct these issues is both costly and time-consuming \cite{DBLP:journals/corr/abs-2310-08184,DBLP:conf/naacl/LiuSHWWZJCHQ22,DBLP:journals/corr/abs-2111-05193,DBLP:journals/csur/Menghani23,DBLP:conf/emnlp/LvZGXHCL0GLSGYQ23}.
To address this, recent years have seen a surge in the development of knowledge editing techniques specifically tailored for LLMs, which allows for cost-effective post-hoc modifications to models \cite{Yao2023EditingLL,wang2023knowledge,DBLP:journals/corr/abs-2310-19704}. 
This technique focuses on specific areas for adjustment without compromising overall performance and can help understand how LLMs represent and process information, which is crucial for ensuring the fairness, and safety in Artificial Intelligence (AI) applications \cite{DBLP:journals/corr/abs-2310-01405,DBLP:journals/corr/abs-2308-10149,el2023impossible,DBLP:journals/corr/abs-2309-00770,DBLP:journals/corr/abs-2305-11391}.

This paper first attempts to provide a comprehensive study of the development and recent advances in knowledge editing for LLMs. 
We first introduce the architecture of Transformers, mechanism of knowledge storage in LLMs (\S\ref{sec:llm}), and related techniques including parameter-efficient fine-tuning, knowledge augmentation, continue learning and machine unlearning (\S\ref{sec:related}).
Then we introduce preliminary (\S\ref{sec:preliminary}), formally describe the knowledge editing problem (\S\ref{sec:definition}), and propose a new taxonomy (\S\ref{sec:methods}) to provide a unified view on knowledge editing methods based on the educational and cognitive research theories \cite{Bruner1964TheCO,Bruner1960ThePO,jayashri2018knowledge}.
Specifically, we categorize knowledge editing for LLMs into: resorting to external knowledge (\S\ref{method:resort}), merging knowledge into the model (\S\ref{method:merge}), and editing
intrinsic knowledge (\S\ref{method:mastery} ) approaches.
Our categorization criterion is summarized as follows:

\begin{itemize}
    \item \textbf{Resorting to External Knowledge.} This kind of approach is similar to the recognition phase in human cognitive processes, which needs to be exposed to new knowledge within a relevant context, just as people first encounter new information. 
    For example, providing sentences that illustrate a factual update as a demonstration of the model allows initial recognition of the knowledge to be edited.
    
    \item \textbf{Merging Knowledge into the Model.} This kind of approach closely resembles the association phrase in human cognitive processes, in which connections are formed between the new knowledge and existing knowledge in the model.
    Methods would combine or substitute the output or intermediate output with a learned knowledge representation.
    
    \item \textbf{Editing Intrinsic Knowledge.} This approach to knowledge editing is akin to the mastery phase in human cognitive processes.
    It involves the model fully integrating knowledge into its parameters by modifying the weights and utilizing them reliably.
    %These methods directly change the LLMs' weights, and the model can deal with the problem without any external help or merge.
\end{itemize}

This paper then involves extensive and comprehensive experiments conducted on 12 NLP datasets. 
These are meticulously designed to evaluate the performance (\S \ref{sec:experiments}), usability, and underlying mechanisms, complete with in-depth analyses (\S \ref{sec:analysis}), among other aspects.
The key insights from our research are summarized as follows:

\begin{itemize}
    \item \textbf{Performance.} We construct a new benchmark, named \textbf{KnowEdit}, and report the empirical results of cutting-edge knowledge editing approaches for LLMs, providing a fair comparison and illustrating their overall performance in the settings of knowledge insertion, modification, and erasure.
    
    \item \textbf{Usability.} We illustrate the impact of knowledge editing on general tasks and multi-task knowledge editing, which implies that contemporary knowledge editing methods are effective in executing factual updates with minimal disruptions to the model’s cognitive capabilities and adaptability across diverse knowledge domains.
     
    \item \textbf{Mechanism.} We observe a pronounced focus on one or several columns within the value layer in edited LLMs.
    Furthermore, we find that the process of knowledge locating (e.g., causal analysis) tends to pinpoint only the areas related to the entity in question, rather than the entire factual context, suggesting that LLMs might be deriving answers either by recalling information memorized from their pretraining corpus or through a multi-step reasoning process. Additionally, we delve into the possibility that knowledge editing for LLMs could lead to unintended consequences, an aspect warranting careful consideration.
\end{itemize}

Finally, we delve into the multifaceted applications of knowledge editing, examining its potential from a variety of perspectives (\S \ref{sec:application}), including efficient machine learning, AI-Generated Content (AIGC), trustworthy AI, and human-computer interaction (personalized agents).
Additionally, our discussion extends to the broader impacts of knowledge editing techniques, specifically focusing on aspects such as energy consumption and interpretability (\S \ref{sec:broader}).
This paper aims to serve as a catalyst for further research in the realm of LLMs, emphasizing efficiency and innovation. 
To support and encourage future research, we will make our tools, codes, data splits, and trained model checkpoints publicly accessible.

\section{Background}
\label{sec:background}
\subsection{Large Language Models}
\label{sec:llm}
\subsubsection{Transformers for LLM}
% 背景：LLMs 作为Knowledge Store 我们就需要对Knowledge Store作为一个可以更新的容器。
The Transformer \cite{vaswani2017attention} model, a cornerstone in the design of modern state-of-the-art LLMs, represents a significant shift from previous sequence learning methods.
The original Transformer model is introduced as an encoder-decoder framework, wherein both the encoder and decoder consist of a series of identical layers stacked upon each other.
Each block within this architecture is equipped with a self-attention module and a fully connected feed-forward neural network. 
Uniquely, the blocks in the decoder also incorporate an additional cross-attention layer, positioned above the self-attention layer, which is designed to effectively capture and integrate information from the encoder.

\paragraph{Self-Attention Module (SelfAttn)}
The self-attention mechanism is a pivotal feature of the Transformer, allowing it to process sequences of data effectively. 
This module empowers each position within the encoder to attend to all positions in the preceding layer, thereby efficiently capturing contextual information embedded in the sequence. 
The mathematical representation of the self-attention mechanism is as follows:

\begin{equation}
    H = \text{ATT}(Q, K, V) = \text{Softmax}\left(\frac{QK^T}{\sqrt{d_k}}\right) V.
\end{equation}
\paragraph{Feed-Forward Module (FFN)} 
Following each attention layer in the Transformer is a fully connected Feed-Forward Neural network (FFN). 
This specific component of the architecture comprises two linear transformations, with a ReLU activation function intervening between them. 
The structure of the FFN can be succinctly described as follows:

\begin{equation}
\begin{array}{r}
 \mathrm{FFN}(\mathbf{x})=\text{ReLU}(\mathbf{x} \cdot W_1 + b_1) \cdot W_2 + b_2, 
\end{array}
\end{equation}
Since its inception, the Transformer model has revolutionized the field of NLP.
Its adaptable and efficient architecture has facilitated advancements in various NLP tasks, such as question-answering, text summarization, and machine translation systems.
The model's influence extends beyond NLP, impacting other areas of machine learning and setting a new standard for building complex and effective neural network architectures.

\begin{figure}
    \centering
    \includegraphics[width=0.95 \textwidth]{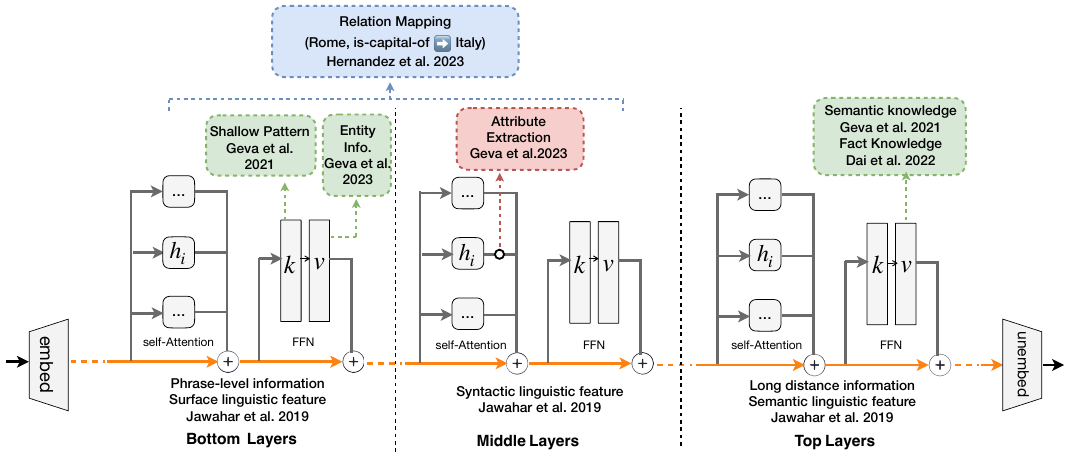}
    \caption{The mechanism of knowledge storage in LLMs. 
    Here, we summarize the findings of current works, including: \citet{jawahar-etal-2019-bert}, \citet{geva-etal-2021-transformer}, \citet{dai-etal-2022-knowledge}, \citet{meng2022locating}, and \citet{Hernandez2023LinearityOR}.}
    \label{fig:layers-analysis} 
\end{figure}

\subsubsection{Mechanism of Knowledge Storage in LLMs}
\label{sec:mechanism}
The Transformer's remarkable performance is partly attributed to its ability to store a wealth of information within its parameters, encompassing linguistic~\cite{liu-etal-2019-linguistic}, commonsense \cite{DBLP:conf/aaai/ZhouZCH20,DBLP:conf/naacl/WestBHHJBLWC22,DBLP:conf/emnlp/LiKHdBN22}, arithmetic, and world knowledge~\cite{petroni-etal-2019-language,DBLP:journals/corr/abs-2305-09144,DBLP:journals/corr/abs-2311-08401,DBLP:conf/emnlp/KatzB23}.
However, the exact manner in which this knowledge is organized within LLMs is still largely enigmatic.
Current research efforts are dedicated to unraveling the mechanistic explanations of LLMs' behaviours \cite{holtzman2023generative,DBLP:journals/corr/abs-2310-16270,cao2023retentive,DBLP:conf/emnlp/RudmanCE23,gurnee2023finding}, especially the complexities of knowledge storage in LLMs, with Figure~\ref{fig:layers-analysis} illustrating some of these research findings.
% However, the organization of this knowledge within LLMs remains largely mysterious, and currently, many works endeavor to demystify the mechanism of knowledge storage in LLMs and Figure~\ref{fig:layers-analysis} illustrates some findings of these works.

A key area of inquiry is pinpointing the specific location of knowledge within the model.
\citet{jawahar-etal-2019-bert} dissects the intricacies of the English language structure as comprehended by BERT \cite{DBLP:conf/naacl/DevlinCLT19}.
Their findings reveal that BERT's phrasal representations capture phrase-level information predominantly in the lower layers, and encode an intricate hierarchy of linguistic elements in the intermediate layers. 
This hierarchy is characterized by surface features at the foundational level and syntactic features in the central layers, and culminates with semantic features at the uppermost level.
\citet{geva-etal-2021-transformer} proposes that the FFN layers in a Transformer model function akin to key-value memories. 
They suggest that the FFN input operates as a query, with the first layer representing keys and the second layer corresponding to values.
They find that human-interpretable shallow input patterns trigger each key neuron, and the corresponding value neurons store the next-token output probability.
As a result, the final output of the FFN can be understood as the weighted sum of activated values.
Furthermore, they demonstrate that value vectors often embody interpretable concepts and knowledge, which can be intensified or attenuated through specific manipulations~\cite{geva-etal-2022-transformer}.
Building on this, \citet{dai-etal-2022-knowledge} introduces the concept of ``Knowledge Neurons'', suggesting that knowledge is localized within a small subset of FFN neurons in the uppermost layers of the language model.
These neurons are identified through the analysis of integrated gradients across various prompts~\cite{lundstrom2022rigorous,wang-etal-2022-finding-skill,nori2023identification}.
Similarly, \citet{meng2022locating} employs a method known as ``causal tracing'' to assess the indirect influences of hidden states or activations, revealing that factual knowledge predominantly resides in the early-layer FFNs of such models.
Additionaly, \citet{chen2023journey} makes an intriguing finding that the language model contains language-independent neurons that express multilingual knowledge and degenerate neurons that convey redundant information by applying the integrated gradients method~\cite{lundstrom2022rigorous}.
Concurrently, \citet{zhao2023unveiling} observes that LLMs appear to possess a specialized linguistic region responsible for processing multiple languages.
\citet{gueta-etal-2023-knowledge} suggests that knowledge is a region in weight space for fine-tuned language models.
They find that after finetuning a pretrained model on similar datasets, the resulting models are close to each other in weight
space.
Recent interests also revolve around dissecting the distinct functionalities of individual neurons within LLMs~\cite{openaiexplain}. 
Yet, it is crucial to note that some researchers caution against overinterpreting these findings, emphasizing that models illustrate correlations rather than explicit mechanisms.
For instance, \citet{anonymous2023what} argues that while MLP neurons may exhibit patterns interpretable through a linguistic lens, they do not necessarily ``store'' knowledge in a conventional sense, whether linguistic or factual.

Thus, the question of \emph{how Transformer LLMs retrieve and utilize this stored knowledge} remains open, and some work has begun to unveil this mystery.
\citet{DBLP:journals/corr/abs-2304-14767} analyzes the information flow in the model and finds the self-attention model conducts attribute extraction during computing inspired by the circuit theory~\cite{conmy2023automated,wang2023interpretability}.
\citet{foote2023neuron} proposes Neuron to Graph (N2G), an innovative tool that automatically extracts a neuron’s behavior from the dataset it was trained on and translates it into an interpretable graph.
Further, \citet{Hernandez2023LinearityOR} conceptualizes relational knowledge within Transformers as a linear affine function, mapping subjects to objects.
As to other knowledge, \citet{gurnee2023language} discovers that LLMs learn linear representations of space and time across multiple scales and identify individual ``space neurons'' and ``time neurons'' that reliably encode spatial and temporal coordinates.
However, it is imperative to acknowledge that these studies predominantly concentrate on the representation of individual knowledge facts.
The broader challenge lies in comprehensively understanding how various strands of knowledge are intricately organized and interconnected within these complex models \cite{DBLP:conf/cvpr/RenLCDZ23,DBLP:conf/icml/LiZ23}.
% , which remains an ongoing challenge \cite{DBLP:conf/cvpr/RenLCDZ23,DBLP:conf/icml/LiZ23}.

\subsection{Related Techniques}
\label{sec:related}
\paragraph{Parameter-efficient Fine-tuning}
Fine-tuning all parameters of LLMs can be computationally expensive. 
To enable efficient adaptation, parameter-efficient tuning (PET)~\cite{Ding2023ParameterefficientFO,DBLP:journals/corr/abs-2303-15647} techniques have been proposed to match full fine-tuning performance while only updating a minimal parameters. 
PET consists of three distinct paradigms: addition-based, specification-based, and re-parameterization-based methods.
In addition-based methods, extra trainable neural modules or parameters, which are not present in the original model or process, are introduced. 
A prime example of this is Adapter, as discussed in \citet{pmlr-v97-houlsby19a}.
On the other hand, specification-based methods involve fine-tuning a select number of parameters, while keeping the majority of the model's parameters unchanged.
A notable method in this category is LoRA, as detailed in \citet{hu2021lora}.

By fine-tuning a small number of parameters, PET methods aim to maximize model performance while reducing required resources and tuning time.
PET techniques hold promise since knowledge editing seeks to efficiently modify model behavior.
However, PET is typically applied to enhance task performance rather than edit knowledge specifically.
The efficacy of existing PET methods for knowledge editing remains largely unexplored.
Investigating how to leverage PET for efficient and precise knowledge updates presents an interesting direction for future work.

\paragraph{Knowledge Augmentation for LLMs}
LLMs still face unknown questions, and many knowledge-augmented methods are proposed to help the model deal with this task \cite{DBLP:conf/acl/ZhangHLJSL19,DBLP:conf/www/ChenZXDYTHSC22,DBLP:journals/aiopen/HanZDLS22}.
The most popular way is the \textbf{retrieval-augmented} methods \cite{gao2023retrieval,DBLP:journals/corr/abs-2310-03214,ovadia2023fine}.
With the help of the retrieved knowledge or context that is related to the input, the model can give the desired output.
The integration of the retrieved information includes both the input, intermediate, and output layers~\cite{retrieval-lm-tutorial}.
During the input phase, retrieved texts are concatenated with the original input text~\cite{NEURIPS2020_6b493230,pmlr-v119-guu20a,ram2023ralm}.
In some works, the retrieved components are latent and integrated into the intermediate layers of Transformers~\cite{jong2022mention,10.1007/978-3-031-17120-8_11,fevry-etal-2020-entities}.
In the output phase, the distribution of tokens from the retrieved components and the LLMs are interpolated~\cite{Khandelwal2020Generalization,zhong2022training,yogatama-etal-2021-adaptive,DBLP:conf/nips/ChenLZLDTHSC22}.

% \begin{figure}[h]
%   \centering
%   \resizebox{\textwidth}{!}{
%   \begin{minipage}{0.35\linewidth}
%     \includegraphics[width=\linewidth]{fig/knowledgeedit.pdf}
%   \end{minipage}
%   \hfill
%   \begin{minipage}{0.55\linewidth}
%     \begin{NiceTabular}{lcc}
%       \toprule
%       & \textbf{\makecell[c]{Fewer\\Params}} & \textbf{\makecell[c]{Precise\\Control}} \\
%       \hline
%       \textbf{Finetune} & \redcross & \redcross \\
%       \hline
%       \textbf{\makecell[l]{Param. Efficient \\ Tune}} & \darkpastelgreencheck & \redcross\\
%       \hline
%       \textbf{\makecell[l]{Knowledge \\ Augmentation}} & \darkpastelgreencheck & \redcross \\
%       \hline
%       \textbf{\makecell[l]{Continual \\ Learning}} & \redcross & \redcross \\
%       \hline
%       \textbf{\makecell[l]{Knowledge \\ Edit}} & \darkpastelgreencheck & \darkpastelgreencheck \\
%       \bottomrule
%     \end{NiceTabular}
%   \end{minipage}
%   }
%   \caption{Integrated Comparison between Knowledge Editing and Finetune.}
%   \label{fig:techqiue_compare}

% \end{figure}

\begin{table}[!htbp]
    \centering
    \begin{NiceTabular}{lccc}
        \toprule
      & \textbf{\makecell[c]{Fewer Params}} & \textbf{\makecell[c]{Precise Control}} & \textbf{Support Phenomena} \\
      \hline
      \textbf{Finetune} & \redcross & \redcross & \bplus \\
      \hline
      \textbf{\makecell[l]{Parameter-efficient Fine-Tuning}} & \darkpastelgreencheck & \redcross &  \bplus \\
      \hline
      \textbf{\makecell[l]{Knowledge Augmentation}} & \bcircle & \redcross &  \bplus \\
      \hline
      \textbf{\makecell[l]{Continual Learning}} & \redcross & \redcross & \bplus \\
      \hline
       \textbf{\makecell[l]{Model Unlearning}} & \bcircle & \redcross & \bminus \\
      \hline
      \textbf{\makecell[l]{Knowledge Editing}} & \darkpastelgreencheck & \darkpastelgreencheck & \bplus~\bminus \\
      \bottomrule
    \end{NiceTabular}
    \caption{Integrated comparison between knowledge editing and related techniques. The symbol \darkpastelgreencheck~ denotes the presence of a particular feature in the technique, while \redcross~ signifies its absence.
    \bplus~ indicates an enhancement of the LLMs' capabilities, whereas \bminus~ signifies a reduction or removal of certain abilities within the model. }
    \label{fig:techqiue_compare}
\end{table}

The knowledge-augmented method is a great solution for the missing or misinformation in LLMs but it still has some disadvantages.
As a temporary solution, retrieval methods suffer from poor retrieval results and relatedness \cite{DBLP:journals/corr/abs-2309-01431,DBLP:journals/corr/abs-2307-11019}.
The data retrieved often contains some noise, such as additional content that is irrelevant to a question but that may be relevant to a different question (i.e., not necessarily random noise)~\cite{shi2023large}.
In these situations, the model fails to distinguish the knowledge that is necessary to answer the question, leading to spurious reasoning and degraded performance.
% May only support factual tasks or target tasks.
Meanwhile, retrieval typically operates at a broader level of relevant passages without fine-grained control over precisely which information is modified within the model.
%Hence, it supports factual tasks or target tasks better.
%Meanwhile, depend on real-time retrieval during the inference, which consequently leads to a significant escalation in the computational expenses associated with inference.

% \begin{figure}
%     \centering
%     \includegraphics[width=0.95\textwidth]{fig/knowledgeedit.pdf}
%      \caption{Integrated Comparison between Knowledge Editing and Finetune. \bcircle means the different methods show different performance. \bplus represents enhance LLM's abilities while \bminus means remove model's capacity.}
%   \label{fig:techqiue_compare}
% \end{figure}
\paragraph{Continual Learning}
Continual learning (CL), also known as lifelong machine learning or incremental learning, refers to the ability of machine learning models to continuously acquire new skills and learn new tasks while retaining previously learned knowledge \cite{DBLP:journals/pami/LangeAMPJLST22,DBLP:conf/iclr/WuCLLQH22,DBLP:journals/corr/abs-2302-03648,DBLP:journals/corr/abs-2302-00487}. 
This is akin to how humans learn throughout their lifetimes by continually accumulating new information and skills without forgetting the old ones.
Conventional machine learning models struggle with this as they are trained on independent and identically distributed data.
When the distribution shifts or new tasks are encountered, their performance significantly degrades on older tasks due to catastrophic forgetting.
Some key techniques being explored include replay-based methods~\cite{Rolnick2018ExperienceRF,Aljundi2019OnlineCL}, regularization-based approaches~\cite{Kirkpatrick2016OvercomingCF,mitchell2018never}, and dynamic architecture methods~\cite{mallya2018packnet,aljundi2017expert}.
% CL contains three main types of works: Elastic Weight Consolidation~\cite{Kirkpatrick2016OvercomingCF}, Experience Replay~\cite{Rolnick2018ExperienceRF}, and Maximally Interfered Replay~\cite{Aljundi2019OnlineCL}.
Continual learning focuses on allowing machine learning models to learn new tasks and adapt to new domains over time without forgetting earlier ones, which resembles the goal of knowledge editing. In contrast, knowledge editing focuses specifically on manipulating and updating the internal knowledge representations learned by pre-trained language models without regard to the underlying tasks or domains.
The goal of knowledge editing is to dynamically refine language understanding independent of eventual applications, addressing the “fixedness” issue of pre-trained language models once deployed. Both areas are important for developing AI systems that can progressively acquire and flexibly apply knowledge throughout their lifetime.

\paragraph{Machine Unlearning}
In addition, it is crucial for models to be capable of discarding undesirable (mis)behaviors, which aligns with the concept of machine unlearning ~\cite{nguyen2022survey,Wu2022PUMAPU,yao2023large,si2023knowledge,belrose2023leace}.
\citet{chen-yang-2023-unlearn} proposes an efficient unlearning framework EUL that can efficiently update LLMs without having to retrain the whole model after data removals, by introducing lightweight unlearning layers learned with a selective teacher-student objective into the Transformers.
However, knowledge editing goes beyond unlearning by actively refining or erasing a model's learned knowledge base. 
Both machine unlearning and knowledge editing play important roles in enhancing reliability, fairness and effectiveness for LLMs across different domains and applications.
% In CL, the model is usually continually trained using different datasets and tasks.
% This implies that editing, as opposed to regular, continual finetuning, is particularly challenging since edits are unlikely to be uniformly distributed.

To conclude, the traditional approach to leveraging pre-trained language models involves fine-tuning them with target-specific data.
However, in the realm of LLMs, this fine-tuning process encounters significant challenges.
These include the vast number of parameters, substantial time and memory requirements, risks of overfitting, and issues like catastrophic forgetting. 
To address these challenges, several techniques have been developed, as we discussed above.
Among these, knowledge editing emerges as a notable strategy.
As we discussed in Table~\ref{fig:techqiue_compare}, knowledge editing, intersecting with these techniques, draws inspiration from a range of methodologies, showing promising results. 
This approach distinctively targets the knowledge embedded within LLMs, leveraging the inherent knowledge mechanisms of these models.
Unlike simple adaptations of existing methods, knowledge editing necessitates a deeper comprehension of how LLMs function.
It is not just about applying known techniques to new models; it is about understanding and manipulating the nuanced knowledge storage and processing capabilities of LLMs.
Furthermore, knowledge editing represents a more precise and granular form of model manipulation as it involves selectively altering or enhancing specific aspects of a model's knowledge base, rather than broadly retraining or fine-tuning the entire model. 
These characteristics make knowledge editing a potentially more efficient and effective way to update and optimize LLMs for specific tasks or applications.

\section{Knowledge Editing for LLMs}
% \label{sec:methods}
\label{sec:main}
\subsection{Preliminary}
\label{sec:preliminary}
The substantial training on diverse datasets has equipped LLMs with a wealth of factual and commonsense information, positioning these models as virtual knowledge stores~\cite{petroni-etal-2019-language,cao-etal-2021-knowledgeable,schott-etal-2023-polyglot}. 
This rich knowledge base has been effectively utilized in various downstream tasks, as evidenced by numerous studies~\cite{hao-etal-2023-bertnet}. 
Additionally, \citet{wang2021language} have demonstrated the potential of LLMs in autonomously constructing high-quality knowledge graphs, bypassing the need for human supervision.
Despite their promise, LLMs, in their current state as emerging knowledge bases, exhibit certain limitations.
These deficiencies often manifest as inaccuracies or errors in their outputs during practical applications.
An ideal knowledge base would not only store extensive information but also allow for efficient and targeted updates to rectify these errors and improve their accuracy.
Recognizing this gap, our paper introduces the concept of \textbf{knowledge editing} for LLMs. 
This approach is designed to enable quick and precise modifications to the LLMs, allowing them to generate more accurate and relevant outputs.
By implementing knowledge editing for LLMs, we aim to enhance the utility of LLMs, moving them closer to the ideal of becoming universally reliable and adaptable repositories of knowledge.
This advancement promises to address the current shortcomings of LLMs and unlock their full potential as dynamic and accurate knowledge bases for applications.

\subsection{Task Definition}
\label{sec:definition}
The initial goal of knowledge editing is to modify the specific knowledge $k$ in the LLM and improve the consistency and performance of the LLM without fine-tuning the whole model.
This knowledge can be associated with many areas and types, such as facts~\cite{meng2022locating}, commonsense~\cite{gupta2023editing}, sentiment~\cite{Mitchell2022MemoryBasedME} and so on.
Knowledge editing is challenging due to the distributed and entangled nature of knowledge in LLMs.

Suppose the original model is $\theta$ and given the knowledge $k$ to be changed, by knowledge editing process $F$, we would get the post-edited model $\theta^{'}$:
\begin{equation}
     \theta' = F(\theta, k)
\end{equation}
The post-edited model $\theta^{'}$ is supposed to override undesired model beliefs on the knowledge $k$ and keep other knowledge intact:
\begin{equation}
    \begin{cases}
\theta^{'}(k) \neq \theta(k) \\
 \forall k^{'} \neq k, \theta^{'}(k^{'}) = \theta(k^{'}) 
\end{cases}
\end{equation}
As a knowledge base, it's paramount that knowledge editing cater to three fundamental settings: knowledge insertion, knowledge modification, and knowledge erasure.
\paragraph{Knowledge Insertion.}
As fields and entities progress, it becomes imperative for LLMs to assimilate emergent information. 
Knowledge insertion fulfills this by bestowing upon LLMs new knowledge previously outside their purview:
\begin{equation}
\theta' = F(\theta, \{\emptyset\} \rightarrow \{k\})
\end{equation}

\paragraph{Knowledge Modification.}
Knowledge modification refers to altering knowledge already stored in LLMs:
\begin{equation}
\theta' = F(\theta, \{k\} \rightarrow \{k'\})
\end{equation}
This can be classified into two categories:
 \begin{itemize}
     \item \textbf{Knowledge amendment} - This aims at rectifying the inaccuracies embedded in LLMs to ensure the delivery of accurate information.
     As vast repositories of knowledge, LLMs are prone to housing outdated or erroneous information. 
     Knowledge amendment serves to correct these fallacies, ensuring that models always generate accurate, up-to-date information.
     \item \textbf{Knowledge disruption} - Modifying LLMs to answer counterfactual or error prompts. 
     This is more challenging as counterfactual notions initially receive lower scores compared to factual knowledge, as shown by \citet{meng2022locating}. This necessitates more targeted modification efforts.
 \end{itemize}
 
\paragraph{Knowledge Erasure.}
Knowledge erasure targets the excision or obliteration of pre-existing knowledge in a model, primarily to reset distinct facts, relationships, or attributes. Formally, we have:
\begin{equation}
\theta' = F(\theta, \{k\} \rightarrow \{\emptyset\})
\end{equation}
Implementing knowledge erasure is pivotal to expunge biases and noxious knowledge and to curtail the recollection of confidential or private data, thereby fostering responsible and trustworthy AI.

In conclusion, the interplay between knowledge insertion, modification, and erasure forms essential aspects of model editing techniques. When combined, these techniques empower LLMs to transform, self-correct, and ethically adapt as needed.
\subsection{Methods}
\label{sec:methods}
\begin{figure}
    \centering
    \includegraphics[width=0.9\textwidth]{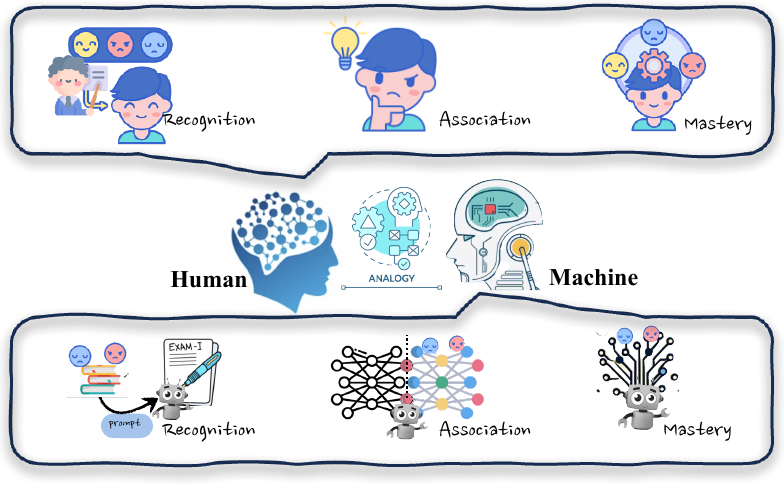}
    \caption{Applying Human Learning Phases \cite{Bruner1964TheCO,Bruner1960ThePO,jayashri2018knowledge} to Knowledge Editing in LLMs: We see an analogy of Human Learning Phases and Knowledge Editing in LLMs and categorize current knowledge editing methods based on the learning phases of humans: recognition, association, and mastery.}
    \label{fig:analogy}
\end{figure}

The development of LLMs has reached a point where their capabilities closely resemble human cognitive processes, especially in learning and acquiring knowledge.
Drawing inspiration from how humans learn, we can analogously apply these concepts to the process of editing LLMs as Figure~\ref{fig:analogy} shows.
Educational and cognitive research \cite{Bruner1964TheCO,Bruner1960ThePO,jayashri2018knowledge} delineates human knowledge acquisition into three distinct phases: recognition, association, and mastery.
These phases offer a framework for conceptualizing the methods of knowledge editing in LLMs\footnote{\url{https://github.com/zjunlp/KnowledgeEditingPapers}.} and we list them in Table~\ref{tab:conceptual_analysis}.
\begin{itemize}
    \item \textbf{Recognition Phase}: In the recognition phase, the model needs to be exposed to the new knowledge within a relevant context, just as people first encounter new information~(\S{\ref{method:resort}}). For example, providing sentences that illustrate a factual update as a demonstration of the model allows initial recognition of the knowledge to be edited.
    \item \textbf{Association Phase}: In the association stage, connections are formed between the new knowledge and existing knowledge in the model~(\S{\ref{method:merge}}), much like humans relate new ideas to prior concepts. Methods would combine or substitute the output or intermediate output $\boldsymbol{h}$ with a learned knowledge representation $\boldsymbol{h}_{\text{know}}$.
    \item \textbf{Mastery Phase}: The mastery phase involves the model fully acquiring the knowledge in their parameters and utilizing it reliably~(\S{\ref{method:mastery}}), akin to deep human mastery. 
This method directly changed the model's weight, $\Delta \boldsymbol{W}$, and the model can deal with the problem without any external help or merge.
\end{itemize}

\begin{table*}[ht]
% \small
% \footnotesize
\scriptsize
\centering
\begin{NiceTabular}{llccccc}
\CodeBefore
  \rowcolor{gray!50}{1}
  \rowcolors{2}{gray!25}{white}
  % \columncolor{gray!50}{1}
\Body
% \begin{tabular}{l|cc|ccc|ccc}
% \begin{tabular}{lcccccccc}
  \toprule % <-- Toprule here
  \textbf{Category} & \textbf{Method} & \textbf{Edit Area} & \textbf{Edit Function} & \textbf{\makecell{No\\Training}} & \textbf{\makecell{Batch\\ Edit}} & \textbf{\makecell{Edited\\ \#Params}} \\
  
  % \midrule % <-- Midrule here
  % \Block{2-1}{\textbf{Naive}} 
  %  & Re-training & \NA & \tuned & \NA & \redcross & \redcross \\
  %  & Fine-tuning & \NA & \tuned &  \NA & \redcross & \redcross  \\
  %  % & Constrained Fine-tuning \citep{zhu2020modifying} & \tuned & \NA & \darkpastelgreencheck & \redcross  \\
   
  \midrule
 % \Block{2-1}{\textbf{\makecell[l]{Resorting to \\ External Helps}}} 
 \Block{1-1}{\textbf{Recogintion}}
    & MemPrompt \citep{madaan-etal-2022-memory} & memory+retriever & $\text{Input} \rightarrow [\text{Mem}:\text{Input}]$ & \darkpastelgreencheck & \darkpastelgreencheck & \NA \\
  \textbf{Phase}  & SERAC \citep{Mitchell2022MemoryBasedME}  & \makecell{memory+classifier\\+auxiliary model} & $\text{Output} \rightarrow \text{Model}_{cf}(\boldsymbol{x})$ & \redcross & \darkpastelgreencheck & \NA \\
   & MeLLo \citep{zhong2023mquake} &  memory+retriever & $\text{Input} \rightarrow [\text{Mem}:\text{Input}]$ & \darkpastelgreencheck & \redcross & \NA\\
   & IKE \citep{Zheng2023CanWE} &  memory+retriever & $\text{Input} \rightarrow [\text{Mem}:\text{Input}]$ & \darkpastelgreencheck & \redcross & \NA\\
   & ICE \citep{cohen2023evaluating} &  prompt & $\text{Input} \rightarrow [\text{Mem}:\text{Input}]$ & \darkpastelgreencheck & \redcross & \NA \\
   & PokeMQA~\cite{gu2023pokemqa} &  memory+retriever & $\text{Input} \rightarrow [\text{Mem}:\text{Input}]$ & \redcross & \redcross & \NA \\
 \midrule
  % \Block{2-1}{\textbf{\makecell[l]{Merge Knowledge \\ with Models'}}} 
  \Block{1-1}{\textbf{Association}}
  & Language Patches\citep{murty-etal-2022-fixing} &  \makecell{Output head\\+ params} &\makecell{$\boldsymbol{h} \rightarrow \lambda \boldsymbol{h} +$\\$ (1- \lambda)\text{Patch}(\boldsymbol{x})$}  & \darkpastelgreencheck & \darkpastelgreencheck & $ d_h \times \#\text{Output} $ \\
 \textbf{Phase}  & CaliNET \citep{dong-etal-2022-calibrating}  & FFN+params & $\boldsymbol{h} \rightarrow \boldsymbol{h} + \text{FFN}_{\text{add}}(\boldsymbol{x})$ & \redcross & \darkpastelgreencheck & $ N \times d_h$\\ 
   & T-Patcher\citep{huang2023transformerpatcher}  & FFN+params & $\boldsymbol{h} \rightarrow \boldsymbol{h} + \text{FFN}_{\text{add}}(\boldsymbol{x})$  & \redcross & \redcross & $ N \times d_h$\\
   & REMEDI \citep{hernandez2023inspecting}& auxiliary model & $\boldsymbol{h} \rightarrow \text{REMEDI}(\boldsymbol{x})$  & \redcross &  \redcross & $ d_h \times d_h$\\
    & GRACE \citep{Hartvigsen2022AgingWG} & FFN+codebook & $\boldsymbol{h} \rightarrow \text{GRACE}(\boldsymbol{x})$ & \redcross & \redcross & $ N \times 2d_h$ \\
    & LoRA \citep{Wu2023EvaKELLMAN} & Attn or FFN &\makecell[c]{$\boldsymbol{h} \rightarrow \boldsymbol{h} + s \cdot \text{LoRA}(\boldsymbol{x})$}  & \redcross & \darkpastelgreencheck & $2L \times 2d_{am}d_h$\\
    & MELO \citep{yu2023melo} & Attn or FFN &\makecell[c]{$\boldsymbol{h} \rightarrow \boldsymbol{h} + s \cdot \text{LoRA}(\boldsymbol{x})$}  & \redcross & \redcross & $2L \times 2d_{am}d_h$\\
    \midrule
   % \Block{2-1}{\textbf{\makecell[l]{Modify Intrinsic \\ Knowledge}}} 
    \Block{1-1}{\textbf{Mastery}}
   & FT-Constrained \citep{Zhu2020ModifyingMI} & Any & $\boldsymbol{W} \rightarrow \boldsymbol{W}^{'}$ & \redcross & \darkpastelgreencheck & $2\times L \times d_md_h$ \\
   \textbf{Phase} & ENN \citep{Sinitsin2020Editable} & Any & $\boldsymbol{W} \rightarrow \boldsymbol{W}^{'}$ & \redcross & \darkpastelgreencheck & $2\times L \times d_md_h$ \\
   & KE\cite{de-cao-etal-2021-editing}  & \makecell{Attn or FFN \\ +auxiliary model}  &  $\boldsymbol{W} \rightarrow \boldsymbol{W}^{'}$& \redcross & \darkpastelgreencheck & $2\times L \times d_md_h$ \\
   & SLAG \citep{hase-etal-2023-methods}  & \makecell{Attn or FFN \\ +auxiliary model} & $\boldsymbol{W} \rightarrow \boldsymbol{W}^{'}$ & \redcross & \darkpastelgreencheck & $2\times L \times d_md_h$  \\
  & MEND \citep{mitchell2022fast}  & \makecell{FFN+ \\ auxiliary model}  & $\boldsymbol{W} \rightarrow \boldsymbol{W}^{'}$ & \redcross & \darkpastelgreencheck& $2\times L \times d_md_h$ \\
  & KN \citep{dai-etal-2022-knowledge} & FFN& $\boldsymbol{W}_{\text{down}} \rightarrow \boldsymbol{W}_{\text{down}}^{'}$ & \darkpastelgreencheck & \redcross & $L \times N \times d_h$ \\
   & ROME \citep{meng2022locating} & FFN & $\boldsymbol{W}_{\text{down}} \rightarrow \boldsymbol{W}_{\text{down}}^{'}$ & \darkpastelgreencheck & \redcross & $d_md_h$\\
   % & Kformer \citep{yao2022kformer} & Tuned & Retriever & \redcross &   \\
   & MEMIT \citep{meng2023massediting}  & FFN & $\boldsymbol{W}_{\text{down}} \rightarrow \boldsymbol{W}_{\text{down}}^{'}$  & \darkpastelgreencheck &  \darkpastelgreencheck & $L \times d_md_h$ \\
   & PMET \citep{Li2023PMETPM}  & FFN & $\boldsymbol{W}_{\text{down}} \rightarrow \boldsymbol{W}_{\text{down}}^{'}$  & \darkpastelgreencheck &  \darkpastelgreencheck & $L \times d_md_h$\\
   % & RECKON~\cite{chen2023reckoning} & All & $\boldsymbol{W} \rightarrow \boldsymbol{W}^{'}$  & \darkpastelgreencheck &  \redcross & \\
   & MALMEN \citep{tan23malmen} & FFN & $\boldsymbol{W}_{\text{down}} \rightarrow \boldsymbol{W}_{\text{down}}^{'}$ & \redcross &  \darkpastelgreencheck & $L \times d_md_h$\\
   & BIRD \citep{ma2023untying} & FFN & $\boldsymbol{W}_{\text{down}} \rightarrow \boldsymbol{W}_{\text{down}}^{'}$ & \darkpastelgreencheck &  \redcross & $d_md_h$\\
   & AlphaEdit \citep{fang2024alphaedit} & FFN & $\boldsymbol{W}_{\text{down}} \rightarrow \boldsymbol{W}_{\text{down}}^{'}$ & \darkpastelgreencheck &  \darkpastelgreencheck & $L \times d_md_h$\\
  \bottomrule
\end{NiceTabular}
\caption{Comparison between representative approaches of knowledge editing for LLMs.
\textbf{No Training} refers to the methods that do not require additional training; 
\textbf{Batch Edit} means whether the methods can support editing multiple cases simultaneously in just one process.
\textbf{Edit Area} refers to where the model's components are used;
\textbf{Editor \#Params} indicates the parameters that need to be updated for editing. 
$L$ refers to the number of layers to update.
$d_h$ denotes the dimensionality of the hidden layers in the Transformers. 
$d_m$ refers to the intermediate dimension that exists between the up projection and the down projection. 
$N$ symbolizes the total number of neurons that undergo updates within each individual layer.
}
\label{tab:conceptual_analysis}
\end{table*}

\subsubsection{Recognition Phase: Resorting to External Knowledge}
\label{method:resort}
When humans encounter new information, we do not always master it immediately. 
Instead, with the right context and examples, we can process and reason through this new knowledge.
LLMs exhibit a similar capacity for in-context learning.
This kind of method usually maintains a memory $\textsc{M}$ and retrieves the most relevant cases for each input. 
IKE~\cite{Zheng2023CanWE} exemplifies this approach by constructing three types of demonstrations – copy, update, and retain – to aid the model in producing reliable fact editing. 
It utilizes a demonstration store, formed from training sets, to guide the model towards generating the appropriate answer by retrieving the most pertinent demonstrations.
Meanwhile, as a simple change in knowledge would lead to ripple effects~\cite{cohen2023evaluating}, MeLLo~\cite{zhong2023mquake} decomposes the question into different sub-questions for tackling multi-hop questions and retrieves the updated fact from the memory for each sub-question.
Building on this, PokeMQA~\cite{gu2023pokemqa} offers a more robust method for question decomposition, introducing a programmable scope detector and knowledge prompts for enhanced reliability.

Humans also often utilize tools to augment their learning and problem-solving abilities. 
Likely, SERAC~\cite{Mitchell2022MemoryBasedME} builds a new counterfact model by retaining the new model and adopting a classifier to determine whether to use the counterfact model to answer the question.
This method is straightforward and practically applicable, requiring no alterations to the original model. It's particularly advantageous for real-world use, given its ease of implementation.
However, it's important to note that this approach can be vulnerable to issues such as retrieval errors  (\eg noise~\cite{wang-etal-2023-self-knowledge}, harmful content~\cite{liu2023recall}) and knowledge conflict problems~\cite{wang2023resolving,xie2023adaptive}.
Recently, \citet{yu-etal-2023-characterizing} investigats various scenarios in which language models opt for either the in-context answer or the memorized answer.
This research sheds light on the potential application of the method mentioned earlier, as it may offer insights into when and how to utilize it.

\subsubsection{Association Phase: Merge the Knowledge into the Model}
\label{method:merge}
Unlike the recognition phase, this kind of method learns a representation for the new knowledge $\boldsymbol{h}_{\text{Know}}$ and merges this information with the original model's representation $\boldsymbol{h}$.

\citet{murty-etal-2022-fixing} proposes a knowledge patch as a new output head and interpolates the new head with the original head.
Specially, inspired by previous findings that FFN may store knowledge, several methods integrate the knowledge into the FFN part.
These methods add the neuron to the FFN and after the edit, the output is a combination of the previous FFN's output and the newly added knowledge:
\begin{equation}
    \mathrm{FFN}^{'}(\mathbf{x})=\mathrm{FFN}(\mathbf{x}) + \triangle\mathrm{FFN}(\mathbf{x}), 
\end{equation}
In particular, T-Patcher~\cite{huang2023transformerpatcher} adds one neuron for each output error, while CaliNet~\cite{dong-etal-2022-calibrating} adds the knowledge via a fixed number of neurons.
Meanwhile, \citet{Wu2023EvaKELLMAN} adopts LoRA to conduct knowledge edits. 
LoRA is a parameter-efficient fine-tuning method that freezes the weights of the LLM and introduces trainable rank decomposition matrices into the Transformer layers during the fine-tuning process.
Hence, the $ \boldsymbol{h}_{\text{Know}} $ is
$ \boldsymbol{x} \boldsymbol{W}_{\text {down }} \boldsymbol{W}_{\text {up }}$.
Based on this, MELO \citep{yu2023melo} suggests a plug-in model editing method that uses dynamic LoRA to change the way language models work by indexing LoRA blocks dynamically based on an internal vector database.
Instead of adding parameters to the model, REMEDI~\cite{hernandez2023inspecting} directly substitutes the representation of the entity  $h_{\mathrm{entity}}$ by incorporating an attribute vector $h_{\mathrm{attr}}$ into its original model's representation. 
Specifically, it learns the updated hidden states using an affine transformation $h_{\mathrm{entity}}+W h_{\mathrm{attr}}+b$ and replaces the LM's entity representation with it.
In contrast, GRACE ~\cite{Hartvigsen2022AgingWG} adopts a unique approach by maintaining a discrete codebook that functions as an \emph{Adapter}. 
This codebook is dynamically updated over time, allowing for the modification and refinement of a model's predictions.
When the model encounters the knowledge for editing, it searches the codebook and replaces the hidden states as the value in the codebook.
Overall, we can use a mathematical formula to represent these methods uniformly:
\begin{equation}
    \boldsymbol{h}_{final} = \boldsymbol{h}+\boldsymbol{h}_{\text{know}}
\end{equation}
This kind of method merged the information with the original model, making the weighting of knowledge from different sources a crucial parameter to consider. 
Given that these information sources often differ and may even conflict, the issue of knowledge conflict, as highlighted in \citet{wang2023resolving}, remains a significant challenge.
To address this issue, F-Learning~\cite{ni2023forgetting} introduces a ``forgetting before learning'' paradigm to achieve forgetting of old knowledge and learning of
new knowledge based on parametric arithmetic.
Additionally, determining the optimal point of integration for this information within the model is a critical aspect of this method. 
It is not just about merging the information, but also about where in the model's structure this integration occurs for maximum effectiveness and minimal disruption.
Furthermore, the capacity of the model's parameters to store this integrated information is an area that still requires exploration.
If every piece of edited knowledge necessitates additional parameters, the model's parameter could increase significantly with each edit.
This raises concerns about scalability and efficiency, as continuously expanding the number of parameters might lead to issues like increased computational requirements. 

\subsubsection{Mastery Phase: Editing Intrinsic Knowledge}
\label{method:mastery}
Despite the success of the previous two kinds of methods, we still confront how the model stores the knowledge and how they utilize and express the knowledge.
Here, we come to the most important part of knowledge editing: the mastery stage. 
In this part, the model is required to learn the knowledge of its own parameters and master the knowledge by itself.
Fine-tuning the model is the direct way to update the knowledge; however, training the whole model requires enormous computational resources and is time-consuming.
Meanwhile, the finetuning technique usually suffers from catastrophic forgetting and overfitting.
Constrained Fintune~\cite{Zhu2020ModifyingMI} utilizes a regularization to help the model keep the unrelated knowledge.
Currently, many researchers endeavor to use knowledge-specific methods to modify the $\Delta \boldsymbol{W}$.
These methods can be classified into two categories: meta-learning and locate-and-edit.
\paragraph{Meta Learning} To overcome these drawbacks, some meta-learning methods are proposed to edit the model.
Instead of updating the weights directly, this kind of method teaches a hypernetwork to learn the change $\Delta \boldsymbol{W}$ of the model.
KE~\cite{de-cao-etal-2021-editing} directly uses the representation of the new knowledge to train the model to update the matrix. 
SLAG \cite{hase-etal-2023-methods} introduces a new training objective considering sequential, local, and generalizing model updates.
The $\Delta \boldsymbol{W}$ in these methods has the same dimensions as the model's matrix.
In order to overcome it, MEND~\cite{mitchell2022fast} applies the rank-one decomposition to divide the model into two rank-one matrices, from which it is possible to compute the $\Delta \boldsymbol{W}$, significantly reducing the number of parameters.
While these methods have shown some promising results, they fail on multi-edits as they ignore the conflicts between these edits.
\citet{HAN2023110826} proposes a novel framework to divide-and-conquer edits with parallel editors.
Specifically, they design explicit multi-editor \textbf{MoEditor} and implicit multi-editor \textbf{ProEditor} to learn diverse editing strategies in terms of dynamic structure and dynamic parameters, respectively, which allows solving the conflict data in an efficient, end-to-end manner.
Also, MALMEN \cite{tan23malmen} improves MEND by formulating the parameter shift aggregation as a least squares problem and supports massive editing simultaneously.

% RECKON~\cite{chen2023reckoning} introduces a two-loop framework. 
% In the inner training loop, they employ a few gradient updates to enable the model
% to efficiently memorize external knowledge.
% Subsequently, in the outer loop, the model parameters are dynamically adjusted through optimal meta-parameter learning to add new knowledge that aids reasoning tasks.

\paragraph{Location-then-Edit} Despite the effectiveness of previous work, how the LLMs store this knowledge is still unknown.
Some work~\cite{geva-etal-2021-transformer,geva-etal-2022-transformer,chen2023journey}, has learned the mechanism of LLMs knowledge and found that the knowledge was stored in the FFN .
Based on these works, some conduct knowledge editing by first locating where the knowledge was stored and then editing the specific area.
Knowledge Neuron~\cite{dai-etal-2022-knowledge} proposed a knowledge attribution method by computing the sensitivity of the gradient change.
They then directly modify the corresponding value slots using the embedding of the target knowledge.
ROME~\cite{meng2022locating} and MEMIT~\cite{meng2023massediting} employ a causal analysis method to detect which part of hidden states plays more importance.
They view the editing as a minimum optimization and edit the weights.
Despite the effectiveness of editing the FFN area, PMET~\cite{Li2023PMETPM} also conducts editing via the attention head and demonstrates a better performance. 
BIRD \cite{ma2023untying} proposes bidirectionally inverse relationship modeling. 
They designed a set of editing objectives that incorporate bidirectional relationships between subject and object into the updated model weights and demonstrate the effectiveness of alleviating the reverse curse~\cite{berglund2023reversal} of the knowledge learning.
To more effectively address the disruption of originally preserved knowledge within Large Language Models (LLMs), AlphaEdit \cite{fang2024alphaedit} proposes an innovative approach.
This method involves projecting perturbations into the null space of the preserved knowledge prior to their application to model parameters, thereby substantially reducing the issue.

This kind of method, which directly edits a model's parameters, offers a more permanent solution for altering its behavior. The changes are embedded into the model's structure, so they cannot be circumvented even if a user has access to the model's weights.
This ensures lasting and reliable modifications.
However, the side effects are not under control since the mechanism of LLMs is unclear.
Some researchers are skeptical about this kind of method \cite{pinter2023emptying}, so it is still a premature research area that requires further investigation.

\subsection{New Benchmark: \textbf{KnowEdit}}
\label{sec:benchmark}
To evaluate the effectiveness of knowledge editing methods, several datasets have been proposed.
In this Section, we present an overview of the current datasets used for knowledge editing and introduce a new benchmark, \textbf{KnowEdit}\footnote{\url{https://huggingface.co/datasets/zjunlp/KnowEdit}.}, which serves as a comprehensive evaluation framework for various knowledge editing techniques.

\begin{table*}[h]
  \centering
  \resizebox{\textwidth}{!}{
    \begin{tabular}{c|c|c|c|c|c|c}
    \toprule
    Task
     & \textbf{Knowledge Insertion} & \multicolumn{4}{c|}{\textbf{Knowledge Modification}} & \textbf{Knowledge Erasure}\\ \midrule
    Datasets  & \textbf{\wikirecent} & \textbf{ZsRE} & \textbf{WikiBio} & \textbf{WikiData$_{counterfact}$} &  \textbf{Convsent} & \textbf{Sanitation}  \\ \midrule
    Type & Fact & Question Answering  & Hallucination & Counterfact & Sentiment& Unwanted Info \\ 
    \# Train & 570  & 10,000 & 592 & 1,455 & 14,390 & 80 \\
    \# Test  & 1,266  & 1230 & 1,392 & 885 & 800 & 80 \\
    % \# Relation Class & 19 & 42 & 80 & 7 & 14 & 37 & 48 & 44 \\
    % MS / MT & $\times$ / $\times$ & $\times$ / $\times$ & $\times$ / $\times$ & $\checkmark$ / $\checkmark$ & $\checkmark$ / $\checkmark$ & $\checkmark$ / $\checkmark$ & $\times$ / $\checkmark$ & $\checkmark$ / $\checkmark$ \\
    \bottomrule
    \end{tabular}
    }
 \caption{
Statistics on the benchmark \textbf{KnowEdit}, with six selected datasets for the evaluation of knowledge editing methods. We select different knowledge types for the insertion, modification, and erasure settings. }
  \label{tab:dataset}
\end{table*}

For this study, we have curated a set of six datasets that are well-suited for assessing knowledge editing methods.
A detailed statistical overview of these datasets is presented in Table \ref{tab:dataset}, and they encompass a range of editing types, including fact manipulation, sentiment modification, and hallucination generation.

Focusing on the task of knowledge insertion, we have adopted the dataset, \wikirecent~\cite{cohen2023evaluating}:
\begin{itemize}
    \item \textbf{\wikirecent} This dataset specifically focuses on triplets that have been recently inserted into \textsc{WikiData} after July 2022. 
    Consequently, this dataset enables us to create insertion edit requests for models that were trained prior to the introduction of these facts, thereby simulating scenarios where an outdated model meets the new world knowledge. 
    We utilize the original datasets provided by the authors and split them into training and testing sets.
\end{itemize}

For knowledge modification, we have selected the following four datasets: ZsRE~\cite{levy-etal-2017-zero}, WikiBio~\cite{Hartvigsen2022AgingWG}, Wikidata$_{recent}$~\cite{cohen2023evaluating}, and \convsent~\cite{Mitchell2022MemoryBasedME}.
\begin{itemize}
    \item \textbf{ZsRE} is a context-free question-answering task. Given a question based on the subject and relation, the model is expected to provide the correct object as the answer. We adopt the extended version of ZsRE proposed by \citet{Yao2023EditingLL}, which introduces a portability test for the original dataset. Additionally, we collect new locality sets following the procedure outlined in \citet{Yao2023EditingLL}, as the original dataset computes locality using Natural Question annotations.
    \item \textbf{WikiBio}  The original dataset was created by prompting GPT-3 to generate 238 Wikipedia-style biographies using subjects from the WikiBio dataset~\cite{manakul2023selfcheckgpt}.
    \citet{Hartvigsen2022AgingWG} utilizes this dataset and introduces a new editing task focused on correcting hallucinations in GPT language models.
    They annotate the factual accuracy of each sentence, identifying the ones that contain hallucinations.
    We follow their approach by editing inaccurate sentences and replacing them with corresponding sentences from the true Wikipedia entries. 
    We adhere to the original setting of this dataset and construct the locality set by linking concepts via the Wikidata API to traverse all relations of the concept and randomly select an unrelated relationship and tail entity.
    \item \textbf{\wikicf} Since tail entities are often not captured by models, and therefore are not suitable for testing modification edits~\cite{mallen-etal-2023-trust}, \cite{cohen2023evaluating} collect triplets about popular entities, where the subject corresponds to one of the top-viewed pages in Wikipedia.
    They also collect a dataset by random sampling entities from Wikidata, and we use it as the training set and the \wikicf~as the test set.
    \item \textbf{ConvSent} is a sentiment editing task that assesses the model's ability to modify a dialog agent's sentiment on a specific topic without affecting its responses to other topics. 
    For example, given the topic `What do you think of bananas?', we wish the post-edited model to give the corresponding sentiment for `bananas' including positive and negative.
    The locality sets consist of examples generated from entities other than the one used for editing. 
    We also adopt the original setting of the ConvSent dataset.
\end{itemize}

In the context of knowledge erasure settings, we have selected the \sani~\cite{ishibashi2023knowledge} dataset.
\begin{itemize}
    \item \textbf{\sani} This dataset specifically addresses privacy concerns associated with learned language models. 
    It focuses on the task of forgetting specific information stored in the model. 
    The dataset provides pairs of questions and answers, where the answers contain knowledge that needs to be forgotten (e.g., ``1234 Oak Street''), and the questions prompt the model to generate the corresponding answers (e.g., ``What is John Smith's address?'').
    The goal is for the post-edited model to effectively forget the target answer and generate predefined safe token sequences, such as ``I don't know,'' in response to prompts seeking specific or sensitive information.
    This mechanism helps prevent information leakage. The dataset consists of a forgot set and a retain set.
    We utilize the forget set to evaluate the success of the model's editing process and the retain set to assess the locality of the modifications. Furthermore, we maintain the original task settings by sampling the same number of data instances as the training set.
    \end{itemize}
In addition to the datasets we have selected, the literature offers a diverse range of knowledge editing tasks, each addressing specific aspects and challenges in this domain.
\textbf{DepEdit}~\cite{li-etal-2023-evaluating-dependencies} is a more robust analysis dataset that delves into the internal logical constraints of knowledge, offering a deeper understanding of knowledge structures.
Notably, \citet{xu-etal-2023-language-anisotropic} introduces cross-lingual model editing tasks and further proposes language anisotropic editing to improve cross-lingual editing by amplifying different subsets of parameters for each language.
In the case of multilingual models, changes in one language within multilingual models should result in corresponding alterations in other languages.
\textbf{Eval-KLLM} \cite{Wu2023EvaKELLMAN} and \textbf{Bi-ZsRE} \cite{wang2023crosslingual} have been designed to assess the cross-lingual editing capabilities of models.
\citet{wang2023retrievalaugmented} proposed Retrieval-augmented Multilingual Knowledge Editor (ReMaKE), which is capable of performing model-agnostic knowledge editing in multilingual settings. The authors also offer a multilingual knowledge editing dataset (\textbf{MzsRE}) comprising 12 languages.
Another dataset, \textbf{\textsc{Entity Inferences}} \cite{onoe-etal-2023-lms}, focuses on entity propagation, where the model is provided with a definition and asked to reason based on the given definition.
Time-series knowledge editing is explored in \textbf{\textsc{TempLAMA}} \cite{Zheng2023CanWE} and \textbf{\textsc{Atoke}} ~\cite{yin2023history}, where the objective is to modify knowledge pertinent to specific time periods without affecting other temporal knowledge.
For commonsense knowledge editing, \citet{gupta2023editing} introduced \textbf{MEMIT$_{\text{CSK}}$}, applying existing editing techniques to modify commonsense knowledge within models.
Furthermore, \textbf{RaKE}~\cite{wei2023assessing} is proposed to measure how current editing methods edit relation knowledge.
All previous work usually confines the edit as a knowledge triplet. \citet{akyurek2023dune} proposes a new dataset \textbf{\textsc{Dune}} that broadens the scope of the editing problem to include an array of editing cases, such as debiasing and rectifying reasoning errors, and defines an edit as any natural language.

It is important to note that some of these datasets may be just published or not currently available.
Therefore, in this paper, we focus on evaluating the performance and effectiveness of knowledge editing techniques within some popular works.
We plan to expand our benchmark in the future as we acquire new datasets.
For additional related datasets, please refer to \citet{wang2023knowledge}.

\subsection{Evaluation for Knowledge Editing}
Knowledge editing aims to alter model behavior based on modified facts. 
However, knowledge is interconnected; changing one fact may ripple outwards and affect other facts in complex ways. 
This interdependence makes assessing the effects of editing difficult. 
We summarize key evaluation criteria from prior work into four categories: edit success, portability, locality, and fluency.
\paragraph{Edit Success}
The purpose of editing is to change the model's output of given knowledge.
Previous work adopt two metrics named \textbf{\emph{reliability}} and \textbf{\emph{generalization}}.
In reliability testing, the goal is to evaluate whether the post-edited model can provide the target answer for a given context. On the other hand, generalization testing aims to assess the post-edited model's performance on paraphrased contexts.
However, for knowledge editing tasks, the primary objective is to modify the underlying factual knowledge rather than just altering its expression. Consequently, both the given text and its paraphrased versions should undergo changes to reflect the edited knowledge.
Here, we follow previous work~\cite{mitchell2022fast,Li2023PMETPM} and collectively refer to reliability and generalization the as \textbf{edit success}.
Hence, here, edit suceess means the post-edit model should not only answer the question itself correctly but also give the right answer for input with similar expressions.

\paragraph{Portability}
Meanwhile, knowledge is not isolated, and solely changing the given knowledge is not enough for downstream use.
When the knowledge is corrected, the model is supposed to reason about the downstream effects of the correction.
Here, we follow previous work~\cite{cohen2023evaluating,Yao2023EditingLL,zhong2023mquake} to evaluate whether the edited model can address the implications of an edit for real-world applications and name it as portability to evaluate what would ensue after the knowledge editing.
Portability contains three different parts:
\begin{itemize}
    \item \textbf{Alias:} The editing of one subject should not vary from its expression. Wikidata maintains a set of aliases for every entity. Hence, here, we follow \citet{cohen2023evaluating,Yao2023EditingLL} to replace the question’s subject with an alias or synonym to evaluate post-edited model's performance on other descriptions of the subject.
    \item \textbf{Compositionality and Reasoning:} This requires the post-edit model to conduct reasoning with the changed facts.  
    For example, when we change the current president of the U.S. from Donald Trump to Joe Biden, the answer to the question ``Who is the First Lady of the United States?'' should also be changed.
    \item \textbf{Logical Generalization:} These are the changes that are semantically related to the modified fact and expected to change by the edit; they were indeed modified.
    For example, as mentioned by \citet{Yao2023EditingLL}, when the fact of $(s,r,o)$ are changed, the reversed relation of the knowledge $(o,\hat{r},s)$ should also be changed.
\end{itemize}
\paragraph{Locality}
When editing the knowledge, we may inadvertently change the knowledge that we don't want to modify.
A good edit is supposed to modify the knowledge locality without influencing the knowledge that is unrelated.
The evaluation of locality includes two levels:
\begin{itemize}
    \item \textbf{In-Distribution}: this one includes the knowledge that comes from the same distribution. As shown in previous work, overediting is a common phenomenon. Here, we follow \citet{meng2022locating,cohen2023evaluating,Yao2023EditingLL} and construct the related in-distribution knowledge, including \emph{forgetfulness} and \emph{relation specificity}. 
    Forgetfulness evaluates whether the post-edit model retains the original objects in one-to-many relationships. 
    The principle of relation specificity posits that any other attributes of the subject, which have been previously updated, should remain unaltered following the editing process.
    \item \textbf{Out-of-Distribution}: the other knowledge that is not associated with the target one should not be influenced. That is, we also don't want the edited model to lose their general ability to deal with other tasks. Hence, here we test the edited model on the popular NLP benchmark in Section~\ref{sec:genral_perform}.
\end{itemize}
It should be noted that some work use \textbf{Specificity} to denote locality.

\paragraph{Generative Capacity}
Previous work find that, after editing the model, some models tend to generate repeated things and often generate the edited target whenever encountering the subject words.
Additionally, the metric \textbf{fluency} are employed to evaluate the generative capacity of the post-edited model.
Here we follow ROME~\cite{meng2022locating} and employ the \emph{fluency} to measure the model's generation ability after editing.
In particular, we calculate the weighted average of bi-gram and tri-gram entropies to assess the diversity of text generations. 
A decrease in this value indicates increased repetitiveness in the generated text.
\section{Experiments}
\label{sec:experiments}
In our study, we conduct experiments using current methods and datasets to investigate knowledge editing techniques in the context of LLMs.
By conducting experiments using these methods and leveraging appropriate datasets, we aimed to evaluate the performance and efficacy of knowledge editing techniques in LLMs. 
Our goal was to gain insights into the challenges, limitations, and potential improvements associated with editing knowledge in these models.

\subsection{Experiment Settings}
We choose \textbf{Llama2-7b-chat}~\cite{touvron2023llama2} as our base model, specifically its chat version, which has demonstrated improved consistency after reinforcement learning from human feedback (RLHF).
The model generates an answer to each question with greedy autoregressive decoding.
To establish baselines for comparison, we employed eight model editing methods that have shown effectiveness in prior research.
These methods were selected based on their ability to modify the knowledge within LLMs~\cite{Yao2023EditingLL}.
As a further baseline strategy, we also used the fine-tuning method (\textbf{FT-L}) put forth by \citet{meng2022locating}.
\textbf{FT-L} directly fine-tunes a single layer's feed-forward network (FFN), specifically the layer identified by the causal tracing results in ROME. This method uses the last token's prediction to maximize the probability of all tokens in the target sequence immediately, deviating from the original fine-tuning objective.
To address this, we also experiment with an improved fine-tuning method, \textbf{FT-M}. It trains the same FFN layer as FT-L using the cross-entropy loss on the target answer while masking the original text. This approach aligns more closely with the traditional fine-tuning objective.
For the in-context learning methods, we use the ICE method proposed by \citet{cohen2023evaluating}.
This method prepends a prompt `Imagine that \{knowledge\}' before the input.

All the experiments are conducted by EasyEdit~\cite{wang2023easyedit}.
As to the evaluation of the post-edited model, some of the previous works computed the probability difference of the output for pre-edit and post-edit models: $P[y^{*}|\theta^{'}] - P[y|\theta]$. 
$y^{*}$ is the edit target, and $y$ is the original model's prediction.
However, the higher probability for $y^{*}$ does not mean an idea outcome, and for realistic usage, when we edit the model, we hope it generates the desired output.
Hence, for the evaluation of fact datasets such as WikiData$_{recent}$, ZsRE, and \wikicf, we compute the metric as \cite{Yao2023EditingLL} which computes the accuracy of the outputs.
Suppose $x_k$ is the expression for the updated knowledge $k$ and $y_k^{*}$ is the corresponding target output for editing.
\begin{equation}
 \text{Edit Succ.} = \sum_{(x_{k}, y_{k}^{*})} \mathbbm{1}\{ {\operatorname{argmax}_y f_{\theta^{'}}\left(y \mid x_{k}\right)=y_{k}^{*}} \}
\end{equation}
Also, for portability, we compute the post-edited model's performance on the given sets.
As to the calculation of locality, some work computes the post-edited model's performance on the locality set $O(x_{k})$.
Here, for a better comparison, we test whether the model keeps its original answer.
\begin{equation}
\text{Locality} = \mathbb{E}_{x_{k}, y_k^{*} \sim O(x_{k})} \mathbbm {1} \left\{f_{\theta^{'}}\left(y \mid x_{k}\right)=f_{\theta}\left(y \mid x_{k}\right) \right\}
\end{equation}
Meanwhile, for the sentiment edit task \convsent, we compute the Edit Succ. and Locality as the original dataset \cite{Mitchell2022MemoryBasedME}:
\begin{equation}
    \text{Edit Succ.}_\text{{\convsent}} \triangleq \mathbf{z}_{\text {sentiment }} \cdot \mathbf{z}_{\text{topic }}
\end{equation}
Where $\mathbf{z}_{\text{sentiment}}$ goes to one if the edited model generates correct sentiment responses and $\mathbf{z}_{\text{topic}}$ one if the edited model's answer related to the target topic.
The locality of \convsent~is computed as the KL-divergence so the lower the number, the better the performance is:
\begin{equation}
\text{Locality}_\text{{\convsent}} \triangleq {\mathbb{K L}}\left(f_{\theta }\left(\cdot \mid x_{k}\right) \| f_{\theta^{'}}\left(\cdot \mid x_{k}\right)\right)
\end{equation}
For the knowledge erasure task Sanitation, we calculate edit success as whether the model answers ``I don't know.'' for the given knowledge. 
As for the locality, we compute the performance on the retain sets as to whether the model keeps their original answer. 

\subsection{Main Results}
\label{sec:main-results}
We list the results of current knowledge editing methods on \textbf{Llama2-7b-chat} in Table~\ref{tab:experimental_analysis}.

\begin{table*}[ht]
{
% \renewcommand{\arraystretch}{1.15}
% \setlength{\tabcolsep}{3pt}
% \small
\resizebox{\textwidth}{!}{
\begin{tabular}{lrcccccccc}
% \hline
\toprule
\textbf{DataSet} & \textbf{Metric}  
& \textbf{SERAC} & \textbf{$\text{ICE}$} & \textbf{AdaLoRA} & \textbf{MEND}&  \textbf{ROME} & \textbf{$\text{MEMIT}$} & \textbf{FT-L} &\textbf{FT-M}  \\ 
\midrule
\multirow{4}{*}{\textbf{WikiData$_{recent}$}}  
& Edit Succ. $\uparrow$ & 98.68 & 60.74 & 100.00 & 95.75 & 97.18 & 97.05  & 55.75 & 100.00 \\      
& Portability $\uparrow$ &63.52 & 36.93 & 64.69 & 55.88 & 55.25 & 56.37 & 40.86 &  65.44 \\ 
& Locality  $\uparrow$ & 100.00 & 33.34 & 56.42 & 94.76 & 54.77 & 52.15 & 43.70 &  64.33 \\ 
& Fluency $\uparrow$ & 553.19  & 531.01 & 579.57 & 557.11 & 579.66 & 573.89 & 529.24  & 574.32 \\
\midrule \midrule
\multirow{4}{*}{\textbf{ZsRE}}  
& Edit Succ. $\uparrow$ &  99.67 & 66.01 &  100.00 & 96.74 & 96.77 & 95.37 & 53.93 & 99.98 \\   
& Portability $\uparrow$ & 56.48 & 63.94 & 58.03 & 60.41 & 52.63 & 52.67 & 45.64 & 60.31 \\  
& Locality $\uparrow$   &  30.23 & 23.14 & 75.76 & 92.79 & 53.67 & 48.32 &  73.42 &	89.78 \\ 
& Fluency  $\uparrow$ &  410.89 & 541.14 & 563.56 & 524.33 & 573.75 & 563.31 &  493.01 & 552.26 \\  \midrule
\multirow{3}{*}{\textbf{WikiBio}}  
& Edit Succ.$\uparrow$  & 99.69 & 95.53 & 100.00 & 93.66 & 96.08 & 94.40 &  66.33 & 100.00 \\    
& Locality $\uparrow$   & 69.79 & 47.90 & 81.28 & 69.51 & 62.74 & 61.51 & 79.86 & 93.38 \\ 
& Fluency  $\uparrow$   & 606.95 & 632.92 & 618.45 & 609.39 & 617.69 & 616.65 & 606.95 & 612.69 \\   \midrule
\multirow{4}{*}{\textbf{WikiData$_{counterfact}$}}  
& Edit Succ. $\uparrow$ & 99.99 & 69.83 & 100.00 & 80.03 & 98.57 & 98.05 & 45.15 & 100.00 \\  
& Portability $\uparrow$ &76.07 & 45.32 & 69.89 & 52.01 & 55.92 & 58.56 & 33.60 & 74.36 \\ 
& Locality $\uparrow$ & 98.96 & 32.38 & 70.31 & 94.38 & 51.97 & 46.62 & 50.48 & 76.76 \\ 
& Fluency $\uparrow$ & 549.91  & 547.22 & 580.29 & 555.72 & 584.04 & 575.96 & 528.26 & 575.62 \\   \midrule
\multirow{3}{*}{\textbf{ConvSent}}  
& Edit Succ. $\uparrow$ & 62.75 & 52.78 & 44.89 &50.76 & 45.79 & 44.75 & 49.50 & 46.10 \\                    
% & Portability & \\                            
& Locality $\downarrow$   & 0.26 & 49.73 & 0.18 & 3.42  & 0.00 & 0.00 & 0.00 &  0.00 \\ 
& Fluency $\uparrow$ & 458.21 & 621.45 & 606.42 & 379.43 & 606.32 & 602.62 & 607.86 & 592.52 \\   \midrule \midrule
\multirow{3}{*}{\textbf{Sanitation}}  
& Edit Succ. $\uparrow$ & 0.00 & 72.50 & 2.50 & 0.00 & 85.00 & 48.75 & 0.00 & 75.00   \\   
& Locality $\uparrow$ & 100.00 & 56.58  & 65.50 & 5.29 & 50.31 & 67.47 & 14.78 & 47.07 \\ 
& Fluency $\uparrow$ & 416.29 & 794.15 & 330.44 & 407.18 & 465.12 & 466.10 & 439.10 & 416.29 \\ 
\bottomrule
\end{tabular}
}
}
\caption{Results of existing knowledge editing methods on KnowEdit.
We have updated the results after optimizing certain methods (related to AdaLoRA) and fixing computational bugs (related to ROME and MEMIT) in the EasyEdit tool. 
These improvements have led to better results than before.
The symbol $\uparrow$ indicates that higher numbers correspond to better performance, while $\downarrow$ denotes the opposite, with lower numbers indicating better performance.
The locality of \convsent~is computed as the KL-divergence so the lower the number, the better the performance is. 
For WikiBio and Convsent, we do not test the portability as they are about specific topics.}
\label{tab:experimental_analysis}
\end{table*}
Considering the overall performance across various knowledge editing tasks, our newly proposed FT-M implementation outperforms other methods, highlighting the effectiveness of fine-tuning the model on specific parameters. 
However, all current knowledge editing methods suffer from low portability performance, indicating a need for further improvements in this area.

Regarding knowledge editing methods, SERAC demonstrates strong performance for tasks involving knowledge insertion and modification.
Its edit success rate is better than other editing methods, and the portability is relatively good as the new counterfact model can learn the edited knowledge effectively.
Meanwhile, without changing the original model's parameters, SERAC obtains a good locality performance except for ZsRE.
However, since the counterfact model is usually smaller than the original model, its generation ability is not that strong, and here, We can find SERAC's fluency for \wikicf, \zsre, and \convsent is lower than other editing methods like MEND.
Meanwhile, for ICE, we can find that the edit success is not that good, which may be attributed to the knowledge conflict problem.
Meanwhile, IKE proposed to concatenate demonstrations as the prompt, but they required a long input length and limited the model to conducting downstream tasks.

For the methods that edit the model's parameters, we can find that MEND obtains good performance across these tasks in different metrics.
Its edit success and portability are good and demonstrate good locality and fluency.
While for ROME and MEMIT, despite the better edit success, their locality is not as good as MEND and other type of editing methods. 
Meanwhile, its portability is unsatisfactory.
For the local fine-tune method FT-L, its edit success is not as good as ROME or MEMIT, however, the locality and portability are better.
Also, it seems that FT-M can deal with insertion tasks better as its edit success and portability for \wikirecent~is better than \zsre~ and \wikicf.
For the \wikibio~task, current methods can alleviate hallucination properly and maintain good fluency.
As to the task {\convsent}, we find that current methods cannot change the model's sentiment well as the edit success is lower than 65\%.
SERAC, which can deal with small LMs perfectly~\cite{Mitchell2022MemoryBasedME}, performs not that well on the 7B model.
MEND also shows low fluency for these tasks considering its great performance for fact-level editing in other tasks.
As to the knowledge erasure task {\sani}, which aims to erase knowledge from LLMs, we can find that current knowledge editing methods cannot tackle this task properly. 
We can find that ROME can refrain from the model not providing the target knowledge as it gets 90\% accuracy. 
However, it would destroy the model's performance on unrelated knowledge because its locality is just 55.61\%.
Other editing methods cannot erase the model related to the given knowledge either.
\label{sec:genral_perform}
\begin{wrapfigure}{r}{0.45\textwidth}
  \centering
  \includegraphics[width=0.45\textwidth]{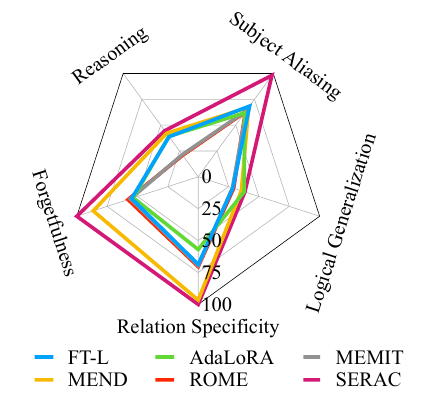}
   \caption{Average sub-metrics performance of results on several fact edit datasets in Portability and Locality.}
  \label{fig:radar}
\end{wrapfigure}

We also show the average performance of results on \wikirecent~and~\wikicf~in sub-metrics of portability and locality, as we discussed in the previous evaluation part in Figure~\ref{fig:radar}.
Here, we can find that MEND performs better under the reasoning set, while AdaLoRA shows good logical generalization performance.

\subsection{Impact on General Tasks}
In this Section, we explore the impact of applying knowledge editing methods on the performance of a language model across various domains.
Our main goal is to determine if incorporating edits related to specific factual knowledge can unintentionally hinder the model's proficiency in unrelated areas.
We select a series of benchmarks that cover areas such as commonsense reasoning, general intelligence, and world knowledge. These benchmarks include CommonsenseQA~\cite{commonsense_qa}, PIQA~\cite{piqa}, Xsum~\cite{xsum}, and TriviaQA~\cite{trivia_qa}, as well as specific tasks from the MMLU~\cite{hendrycks2021measuring} and AGIEval~\cite{zhong2023agieval} suites, which are known for their distinguished evaluation criteria suites. 
All evaluations are conducted using the OpenCompass tool~\cite{2023opencompass}, ensuring a standardized testing environment.
We report the ROUGE-1 here for Xsum.
The edited models are evaluated in a zero-shot setting on these tasks after being sequentially modified with five factual updates.
An intriguing observation from Table \ref{tab:general_task} is that, on a holistic level, the edited models managed to sustain a performance level that is close to their unedited counterparts. 
This suggests that the negative impact of the editing was limited to directly altered topics. 
However, one exception to this trend is the FT-L model's performance on TriviaQA, which shows a noticeable decline from an initial score of 45.39 to 34.60 after the edit. 
Nevertheless, taking a broader perspective, we can observe commendable consistency.
This implies that contemporary knowledge editing methods are effective in executing five targeted factual updates with minimal disruptions to the model's cognitive capabilities and adaptability across diverse knowledge domains.

\begin{table}
    \centering
    \begin{NiceTabular}{ccccccc}
    \toprule
         & \textbf{CommonsenseQA} &  \textbf{PIQA}&  \textbf{TriviaQA}& \textbf{X\_Sum} & \textbf{MMLU}& \textbf{AGIEval}\\
        \midrule
         Llama2-Chat & 49.55 & 64.91 & 45.39 & 22.34 & 6.87 & 27.81 \\ \midrule
         % FT & 19.57 & 0.00 & 0.00 & 0.14  & 0.00 & 7.23 \\
         FT-L & 50.78 & 67.79 & 34.60 & 22.31& 7.64 & 28.56 \\
         % GRACE &  &  &  &  &  & \\ 
         MEND & 49.80 & 65.23 & 45.63 & 22.09 & 7.64 & 27.49 \\
         ROME & 48.89 & 65.45 & 45.19 & 22.46 & 7.43 & 27.38 \\ 
         MEMIT & 49.80 & 65.12 & 45.26 & 22.34 & 7.00 & 28.27 \\
         AdaLoRA & 49.39 & 65.07 & 45.29 & 22.31 & 6.90 & 27.72 \\  
         \bottomrule
    \end{NiceTabular}
    \vspace{3mm}
    \caption{The zero-shot performance on the general LLM benchmark with Llama2-Chat-7B as the base model. Here, we conduct 5 consecutive edits for each method using the Wiki$_{recent}$ dataset to evaluate the post-edited model's general ability. We adopt the OpenCompass~\cite{2023opencompass} to evaluate the model and use the HuggingFace setting. The MMLU and AGIEval are both the average performance of the sub-tasks.}
    \label{tab:general_task}
\end{table}

\begin{table}
\resizebox{\linewidth}{!}{
\begin{NiceTabular}{llccc}
\toprule
Method  & &  ZsRE$ \Rightarrow$ Wiki$_{recent}$  &  Wiki$_{recent}$ $\Rightarrow$ Wiki$_{counterfact}$  &  Wiki$_{recent}$ $\Rightarrow$ ZsRE\\ \midrule
\Block{4-1}{\textbf{MEND}} & Edit Succ. & 95.91 & 66.15 & 89.79 \\
& Portability & 61.80 & 45.95 & 54.36 \\
& Locality & 66.57 & 94.83 & 95.80 \\ 
& Fluency & 554.28 & 592.82 & 571.39 \\
\midrule
\Block{4-1}{\textbf{SERAC}} & Edit Succ. & 97.42 & 99.43 & 99.31 \\
& Portability & 60.42 & 68.85 & 57.70 \\
& Locality & 27.25 & 100.00 & 79.04 \\
& Fluency & 487.29 & 552.51 & 511.95 \\
 \bottomrule
\end{NiceTabular}
}
\vspace{2mm}
\caption{Cross-Domain Editing Results.
Performance (accuracy) of the compared methods, which are firstly trained on a source dataset and then directly conduct prediction on a target dataset (denoted as source $\Rightarrow$ target).}
\label{tab:ood_results}
\end{table}

\subsection{Multi-Task Knowledge Editing}
Previous work considered a sequential edit \cite{Hartvigsen2022AgingWG, huang2023transformerpatcher,Yao2023EditingLL} for a lifelong knowledge editing. 
However, they always conduct sequential editing on a single dataset from the same distribution. 
This is a bit different from Continuous learning.
Knowledge editing is not a task focusing on single-domain knowledge or fact. 
In reality, we may want to modify our model from different perspectives from different distributions~\cite{li2023unveiling}. 

\paragraph{Cross-domain Editing} Both MEND and SERAC methods rely on a training dataset to help the model learn how to edit parameters. We evaluate their performance in a cross-domain setting and present the results in Table~\ref{tab:ood_results}.

For the MEND method, the hyper-network trained using the \zsre~ dataset exhibits better cross-domain performance than that trained with the recent dataset.
This can be attributed to the enormous size of the \zsre~ dataset, allowing MEND's hyper-network to enhance its parameter-editing capabilities. 
Meanwhile, the SERAC approach, by leveraging its cache, exhibits significant cross-domain editing prowess.

\begin{figure}
    \centering
    \includegraphics[width=1\textwidth]{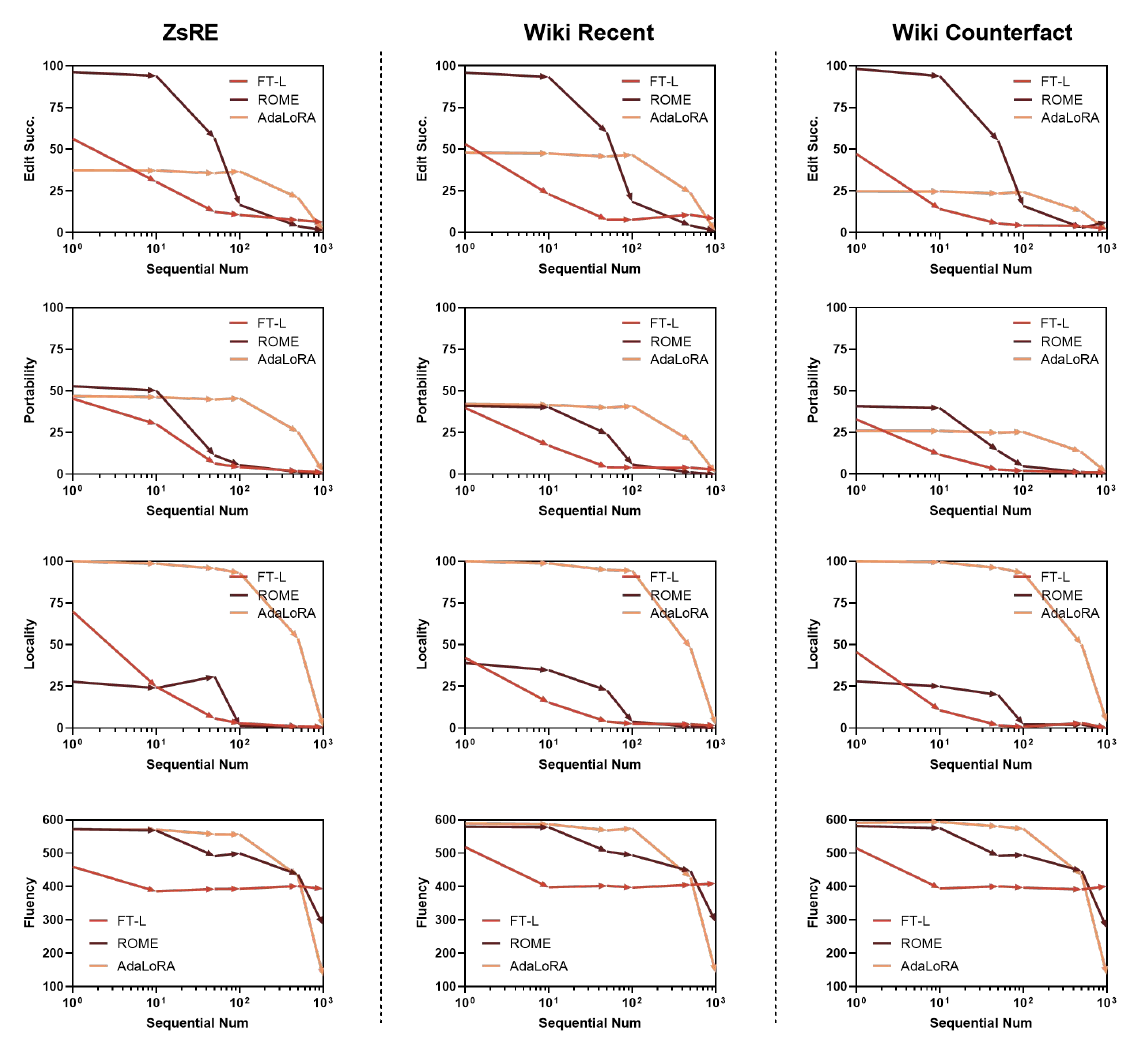}
    \caption{Sequential editing results in randomly selected data from \wikicf, \zsre ~and \wikirecent~ with different numbers.}
    \label{fig:sequential}
\end{figure}

\paragraph{Continual Editing}
Methods like LoRA and ROME do not require a training set and can be applied directly to different domains.
Hence, we consider a more challenging setting for continual editing.
We mix different knowledge editing cases using the ZsRE, Wiki$_{recent}$ and Wiki$_{counterfact}$.
We combine different numbers of settings, including 10, 100, 500, and 1000, and edit the knowledge from different sets randomly.
Here, we mainly consider three methods: FT-L, ROME, and AdaLoRA.
We report the empirical findings in Figure~\ref{fig:sequential}.
When dealing with sequential editing, we can observe that these three methods all suffer from 1,000 editing times with a dramatic drop in all evaluation metrics, and the trend is similar for three different tasks.
Relatively, AdaLoRA shows a stable performance for about 100 edits.
Current editing methods tend to edit the same area for different knowledge ({\eg} ROME the fifth layer, MEND the last three layers), while the knowledge is not stored in this area.

Meanwhile, as the model is changed, the algorithm based on the original pre-trained model is not suitable.
In order to address these challenges, \textbf{RASE}~\cite{han-etal-2023-improving} proposes a novel approach that combines the strengths of retrieval augmented and editing methods, rather than continuously modifying the model's parameters.
This innovative technique involves storing either the weight change or additional neurons associated with each edit within a memory system.
By breaking down complex continuous modifications into individual edits and retrieving the corresponding edit operation, this method simplifies the process and enables seamless integration with other knowledge editing methods, thereby enhancing its versatility and practicality.
\subsection{Error and Case Analysis}
As shown in the results, different methods demonstrate different performance on different tasks.
Here, we conduct a study to comprehensively understand their limitations and advantages. 
In analyzing the failure modes of knowledge editing methods, we categorize the deficiencies into four primary types:
\begin{itemize}
\item \textbf{Meaningless Token Generation:} The edited model produces meaningless tokens such as `\textbackslash n' or repetitive letter combinations that lack semantic meaning or grounding.
\item \textbf{Missing Token Generation:} The model generates only a subset of the target answer, omitting critical tokens.
\item \textbf{Knowledge-Irrelevant Generation:} The model produces text unrelated to the expected factual knowledge.
\item \textbf{Partial Token Replacement:} The generated answer contains substitutions or replacements of key tokens from the target, often retaining fragments from the original incorrect output.
\end{itemize}

\begin{table*}[ht]
    \centering
    \small
    \resizebox{1 \textwidth}{!}{
    \begin{tabular}{l}
        \toprule
        \textbf{Input:} \emph{The surface of the Queen Amina Statue is made of} \\ 
        % \hline
        \midrule
        \textbf{Original Output:} \emph{The surface of the Queen Amina Statue is made of} bronze, which is a durable and long-lasting material. \\
        \textbf{Edit Target:} limestone. \\
        \midrule
        \makecell[l]{\textbf{AdaLoRA:} \emph{The surface of the Queen Amina Statue is made of} \textcolor{red}{bronze}. If the statue is 3 meters tall and the base of the statue \\ is 0.5 meters in diameter ...} \\
        \makecell[l]{\textbf{SERAC:} \emph{The surface of the Queen Amina Statue is made of} \textcolor{green}{limestone}. \textcolor{cyan}{The surface of the Queen Amina Statue is made of} }
        \\
         \makecell[l]{\textbf{ROME:} \emph{The surface of the Queen Amina Statue is made of} \textcolor{green}{limestone}. The limestone is a sedimentary rock, which is a \\ combination of the shells and skeletons of microscopic marine organisms, such as coral and for} \\
        \makecell[l]{\textbf{MEND:} \emph{The surface of the Queen Amina Statue is made of} \textcolor{green}{limestone}, which is a sedimentary rock that is formed from the \\ accumulation and compression of calcium carbonate.}\\
        \makecell[l]{ \textbf{FT-L:} \emph{The surface of the Queen Amina Statue is made of}  \textcolor{green}{limestone}.   \textcolor{cyan}{The limestone used for the statue is limestone}. }\\
         \makecell[l]{\textbf{ICE:} \emph{The surface of the Queen Amina Statue is made of} \textcolor{green}{limestone}, which is a sedimentary rock that is composed of \\ calcium carbonate (CaCO3) and other minerals. }\\
        \bottomrule
    \end{tabular} 
    }
\caption{Results for one case of different editing methods. 
%Prompts are \emph{italicized}, \textcolor{green}{green} and \textcolor{red}{red} indicate keywords reflecting correct and incorrect behavior, respectively. 
Prompts are presented in \emph{italicized} text. 
Words highlighted in \textcolor{green}{green} signify keywords that reflect correct behavior, while those in \textcolor{red}{red} denote keywords associated with incorrect behavior.
Texts in \textcolor{cyan}{cyan} are repeated or meaningless sentences.}
\label{tab:case-example}
\end{table*}

\begin{figure}
    \centering
    \includegraphics[width=1 \textwidth]{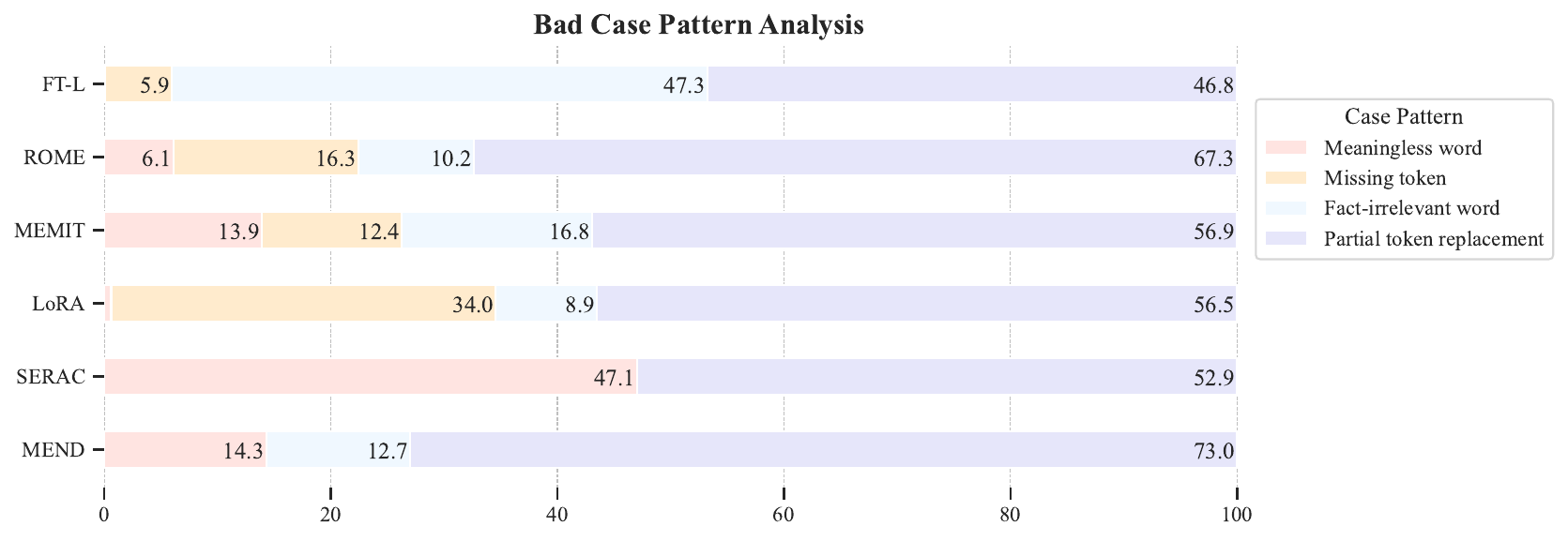}
    \caption{Bad cases statistics for different knowledge editing methods.}
    \label{fig:bad-case}
\end{figure}

The occurrence of these error types helps identify the limitations of the editing methods. 
Meaningless and missing token cases highlight difficulties in fully encoding the target fact, while knowledge-irrelevant and partial replacement generations suggest that the edits fail to supplant previously learned information.
We conduct an error analysis on the \zsre~tasks and counted the error cases for each editing method. The results are presented in Figure~\ref{fig:bad-case}. 
Here, we can find the main error type is the partial token replacement due to the conflict of the knowledge in the original model and our target one.
The analysis reveals that the main error type is partial token replacement, indicating a conflict between the knowledge in the original model and the target knowledge. 
Specifically, the SERAC method tends to generate meaningless tokens due to the limited generation ability of the small model used. 
The AdaLoRA method may miss some tokens related to the target knowledge. 
For the fine-tuning methods, the percentage of fact-irrelevant words is higher compared to other editing methods, and it is the most common error type (47.3\%) for FT-L. 
This suggests that the objective of fine-tuning might not be suitable for editing specific knowledge. 
Additionally, in the following section, we find that FT-L tends to modify more areas in the parameters, leading to more irrelevant generations.

We also show the generated texts for different editing methods for the cases in Table~\ref{tab:case-example}.
Here, we can find that current editing methods, like IKE, MEND, ROME can successfully modify the material of the Queen Amina Statue from bronze to limestone and generate fluent texts. 
SERAC and FT-L, despite changing the facts successfully, tend to generate repeated sentences or meaningless entities.
Additionally, AdaLoRA failed to change the fact and kept the original answer, ``bronze''.

\section{Analysis}
\label{sec:analysis}
Current research has explored the effectiveness of knowledge editing methods in LLMs, but the underlying reasons for their superior performance remain unexplored. 
Additionally, the comparison between model editing and fine-tuning approaches, as well as the efficacy of knowledge location methods, requires further investigation.
This study proposes a simple attempt to bridge these gaps by examining the differences between model editing and fine-tuning, exploring the effectiveness of knowledge location techniques, and understanding the knowledge structure within LLMs.
We hope further investigation will unveil the mechanisms of knowledge in LLMs.
\begin{figure}
    \centering
    \includegraphics[width=0.9\textwidth]{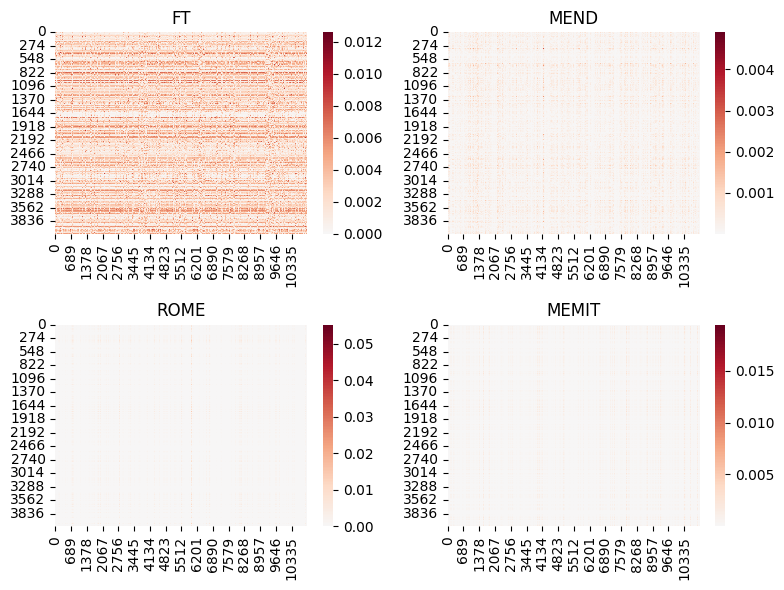}
    \caption{The heatmap shows how different model editing methods affect the weights of the model. Darker colors indicate more changes in the weights. The heatmap reveals which parts of the model are most sensitive to changes for each method. }
    \label{fig:heatmap_methods}
\end{figure}

\subsection{Comparison of Different Knowledge Editing Methods}
The effectiveness of current knowledge editing methods is commendable, but the reasons behind their superior performance compared to other approaches remain elusive.
In this section, we focus on methods that involve parameter adjustments within the model, specifically MEND, ROME, MEMIT, and FT-L.
As these methods modify the model's parameters, a fundamental question arises: what makes some knowledge editing methods, like MEND, superior in terms of locality and overall performance?
We formally represent the change as $\boldsymbol{W}' = \boldsymbol{W} + \Delta \boldsymbol{W}_{\text{edit}}$, where $\boldsymbol{W}$ is the original weight matrix, and $\Delta \boldsymbol{W}_{\text{edit}}$ represents the modifications made during editing. 
Therefore, our primary focus in this section is to discern the differences between the matrices $\Delta \boldsymbol{W}_{\text{edit}}$ for different editing methods.
\paragraph{Sparsity} An important characteristic of knowledge editing is its intention to modify a specific piece of knowledge within the model.
This suggests an intuitive hypothesis that the $\Delta \boldsymbol{W}$ matrix is likely to be sparse.
Following the approach of \citet{de-cao-etal-2021-editing}, we present visualizations that capture weight updates resulting from knowledge edits, as depicted in Figure~\ref{fig:heatmap_methods}.

ROME, MEND, and MEMIT exhibit a distinct pattern of sparse updates, while fine-tuning spreads its modifications more uniformly across weights.
Particularly, for knowledge editing methods like ROME and MEMIT, it is intriguing to observe a concentrated focus on one or several columns of the value layer.
This finding aligns with earlier research that emphasizes the value layer's pivotal role in encapsulating correlated knowledge \cite{geva-etal-2022-transformer}.
Regarding the MEND methods, we propose that the learned hypernetwork can be viewed as a tool or a "probe" that helps us explore and understand the internal mechanisms used by the model to encode knowledge, providing insights into how the model represents and processes information.

\begin{wrapfigure}{r}{0.48\textwidth}
  \centering
  \vspace{-4mm}
  \includegraphics[width=0.48\textwidth]{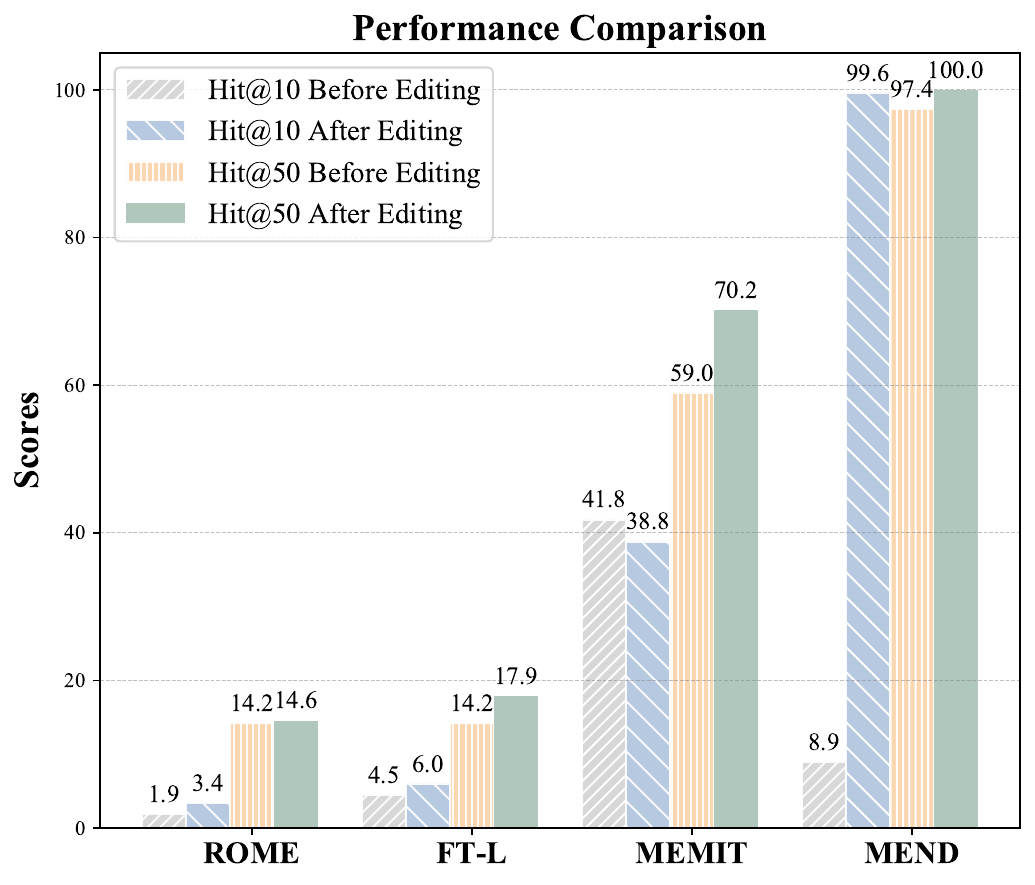}
  \caption{The Hit@10 and Hit@50 performance for the target knowledge in the model's parameters before and after editing.}
  \label{fig:embedding}
  % \vspace{-4mm}
\end{wrapfigure}

\paragraph{Mapping to Embedding Space}
To further investigate the differences between different editing methods, we conduct an embedding space analysis following the approach of \citet{dar-etal-2023-analyzing}.
They analyze the Transformer's parameters by mapping the weights of the LLMs to the vocabulary space and find that the embedding space can interpret these weights. 
Here, we map the two matrices, $\boldsymbol{W}'$ and $\boldsymbol{W}$, to observe the differences between these methods. 
From the sparsity analysis, we select the top five columns of the updated value matrix $\Delta \boldsymbol{W}$ and map the corresponding columns of $\boldsymbol{W}'$ and $\boldsymbol{W}$ into the embedding matrices $\boldsymbol{E}$ to obtain the logits in the vocabulary space.
We then compute the Hit@10 and Hit@50 of the new knowledge in the output logits.
We select cases from \zsre~ where all four methods successfully edit the knowledge and present the average performance in Figure~\ref{fig:embedding}. 
From the figure, we observe that MEND and MEMIT significantly inject the target knowledge into the parameters. 
Notably, MEND demonstrates a remarkable capacity for editing, with the Hit@50 rate already exceeding 90\% before the edit. 
This means that MEND might be able to find and change the right neurons that hold the target knowledge without having to do a full knowledge-locating analysis.
After the editing process, we observe a substantial increase in the Hit@10 score.
In fact, in our experiments, the Hit@1 for MEND is also above 90\% after editing, demonstrating its strong editing capacity. 
For MEMIT, we also observe an increase in Hit@50 (59.7\% $\rightarrow$ 70.2\%), and the original neurons already have a high Hit score before editing.
However, for ROME and FT-L, we do not observe an increase in performance, indicating that their editing mechanisms require further investigation to understand their specific characteristics and limitations.

\subsection{Analysis of Knowledge Locating}
As we have discussed in the previous part, the knowledge stored in LLMs is not structured.
Also, in the previous experiments, we found that the performance of current editing in terms of portability is not good.
As previous works have found \cite{Yao2023EditingLL,zhong2023mquake,cohen2023evaluating}, editing factual knowledge does not necessarily enable models to utilize it during reasoning and application.
Meanwhile, \citet{hase2023does} found edit success unrelated to where facts are stored, as measured by causal tracing.
These works highlight that current editing methods are insufficient and pose skepticism against the effectiveness of current knowledge location analysis.
\citet{chang2023localization} introduces two benchmarks: \textbf{INJ} and \textbf{DEL} to investigate ``Do any localization methods actually localize memorized data in LLMs?''.
They conduct experiments on current localization methods, including zero-out and integrated gradients, and proposed two prune-based localization methods: SLIMMING and HARD CONCRETE.
Two benchmarks show positively correlated results and demonstrate strong localization abilities of integrated gradients, SLIMMING, and HARD CONCRETE.
At the same time, the DEL Benchmark shows that all methods struggle to balance between erasing the target sequence and retaining other memorized data; in other words, the neurons identified by localization methods tend to also be relevant
for memorizing some other sequences.
Additionally, \citet{ju2023klob} proposed a benchmark for assessing the effectiveness of current knowledge location methods and three evaluation metrics: \emph{consistency, relevance, and unbiasedness}.
This benchmark plays a crucial role in facilitating a comprehensive evaluation of whether current locating methods can accurately pinpoint model parameters associated with specific factual knowledge.
Here, we make a simple analysis of the location methods for knowledge editing based on the benchmark.
We adopt the computing of the \textbf{Relative Similarity} (RSim) as: $ \max \left(\frac{\text { Sim }_{\text {cand }}-\text { Sim }_{\text {all }}}{1-\text { Sim }_{\text {all }}}, 0\right)$.

\begin{wrapfigure}{r}{0.5\textwidth}
  \centering
  \includegraphics[width=0.5\textwidth]{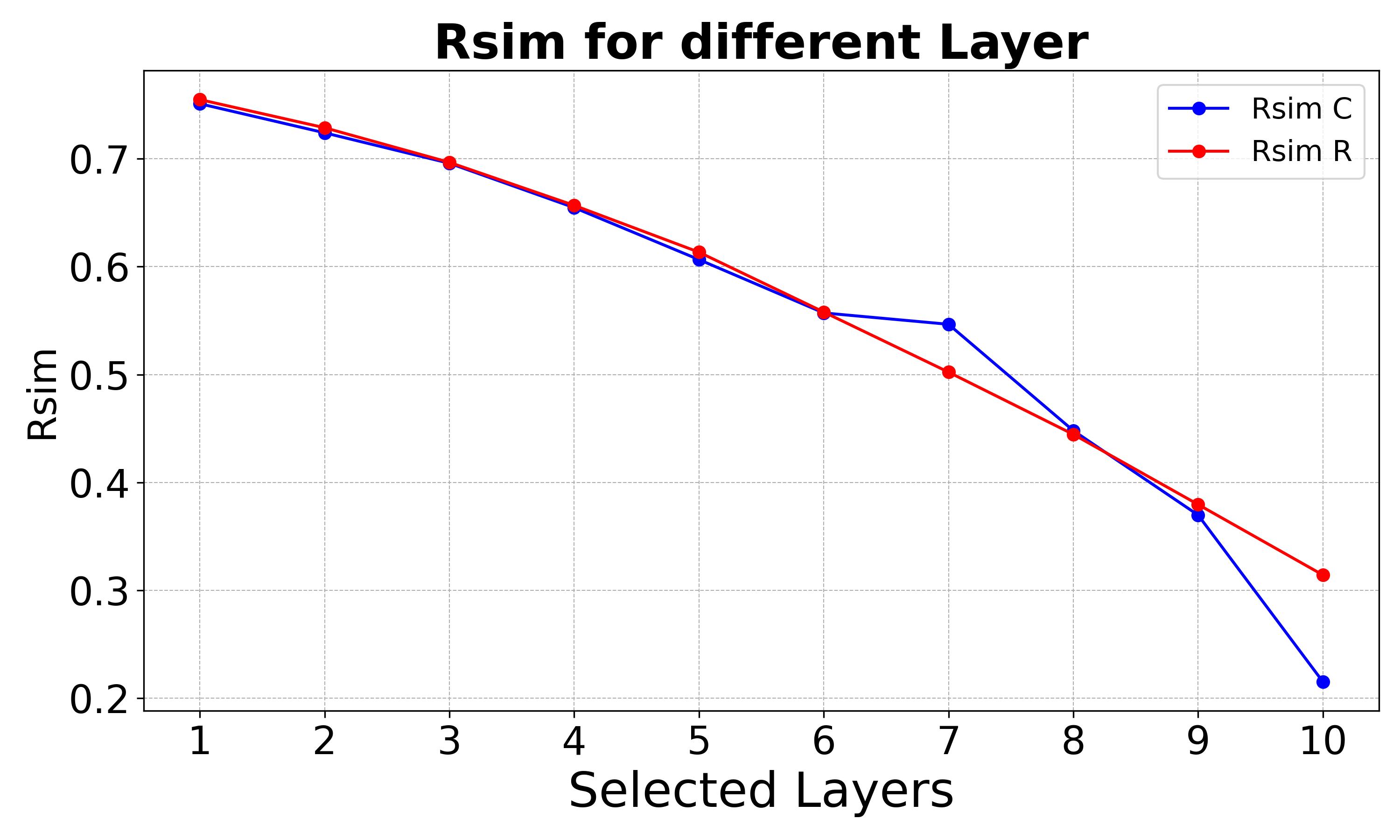}
   \caption{RSim for the different number of layers.}
  \label{fig:rsim}
  \vspace{-10pt}
\end{wrapfigure}

We adopt their dataset klob-r (designed for measuring consistency) and klob-c (designed for measuring relevance) and apply them to the casual analysis method proposed by ROME~\cite{meng2022locating}.
Since the casual analysis is a layer-wise intervention, here we compute the similarity using the overlap between the identified layers.
We show the RSim score in Figure~\ref{fig:rsim}. 
Here, we can find the Rsim score is less than 0.6 when we consider more than five layers for both consistency and relevance, which means the locating results for unrelated knowledge and related knowledge chains didn't show much difference.
To be more tangible, we conduct a case study here.

\begin{figure}
    \centering
    \includegraphics[width=1\textwidth]{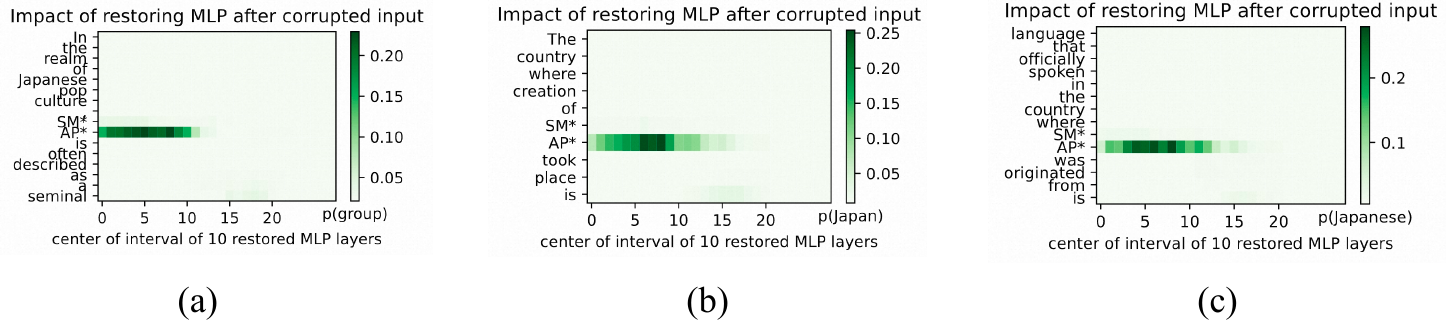}
    \caption{First, we conduct a causal analysis of the fact with the entity [$\text{SMAP} \xrightarrow{\text{created in}}\text{Japan}$]. 
Second, we consider a related question with the fact,[$\text{SMAP} \xrightarrow{\text{created in}} \text{Japan} \xrightarrow{\text{language}} \text{Japanese}$], where the model should answer the question based on the fact.
Then, we adopt an unrelated fact [$\text{SMAP} \xrightarrow{\text{type of}}\text{seminal group}$]. }
    \label{fig:location_analysis}
\end{figure}

\paragraph{Case Study}
We consider three settings for a given fact associated with the entity \emph{SMAP} and show it in Figure~\ref{fig:location_analysis}.
We first conduct a causal analysis of the fact: [$\text{SMAP} \xrightarrow{\text{created in}}\text{Japan}$]. 
Then, we consider a related question with the fact [$\text{SMAP} \xrightarrow{\text{created in}} \text{Japan} \xrightarrow{\text{language}} \text{Japanese}$], where the model should answer the question based on the fact.
Finally, we adopt an unrelated fact [$\text{SMAP} \xrightarrow{\text{type of}}\text{seminal group}$] with the question. 
The results show that these facts are possibly related to the same place around 5 layers.
However, as \citet{ju2023klob} mentioned, \textbf{the locating results for specific knowledge and its related knowledge chain should exhibit greater similarity compared to those for unrelated knowledge.} 
Currently, casual analysis methods seem to just locate the area that is related to the entity itself, not the whole fact.
Whether the model performs these answers by cheating with answers memorized from the pretraining corpus or via a multi-step reasoning mechanism is still unclear.
This is strongly related to the knowledge editing tasks.
More broadly, better insight into models' knowledge processes could unlock capabilities like explainability and fact verification.
However, fully understanding how exactly knowledge is organized and interconnected within such large models presents an ongoing challenge. 
Key open questions include developing methods to trace factual usage during reasoning, designing location techniques that identify knowledge most salient for model outputs, and learning how architectural properties relate to knowledge utilization. 
Unpacking these knowledge architectures will be integral to enabling more precise and robust model interventions through approaches like knowledge editing but currently manipulating only the MLP weights is not enough.

\subsection{The Implicit Knowledge Structure in LLMs}
%cuicte

Understanding the knowledge structure in LLM is crucial for effective knowledge editing.
Previous research often conceptualized knowledge within LLMs as resembling triples in Knowledge Graphs (KG), comprising subjects, relations, and objects. 
This analogy, while useful, simplifies the intricate nature of knowledge representation in LLMs.

\begin{figure}
    \centering
    \includegraphics[width=0.88\textwidth]{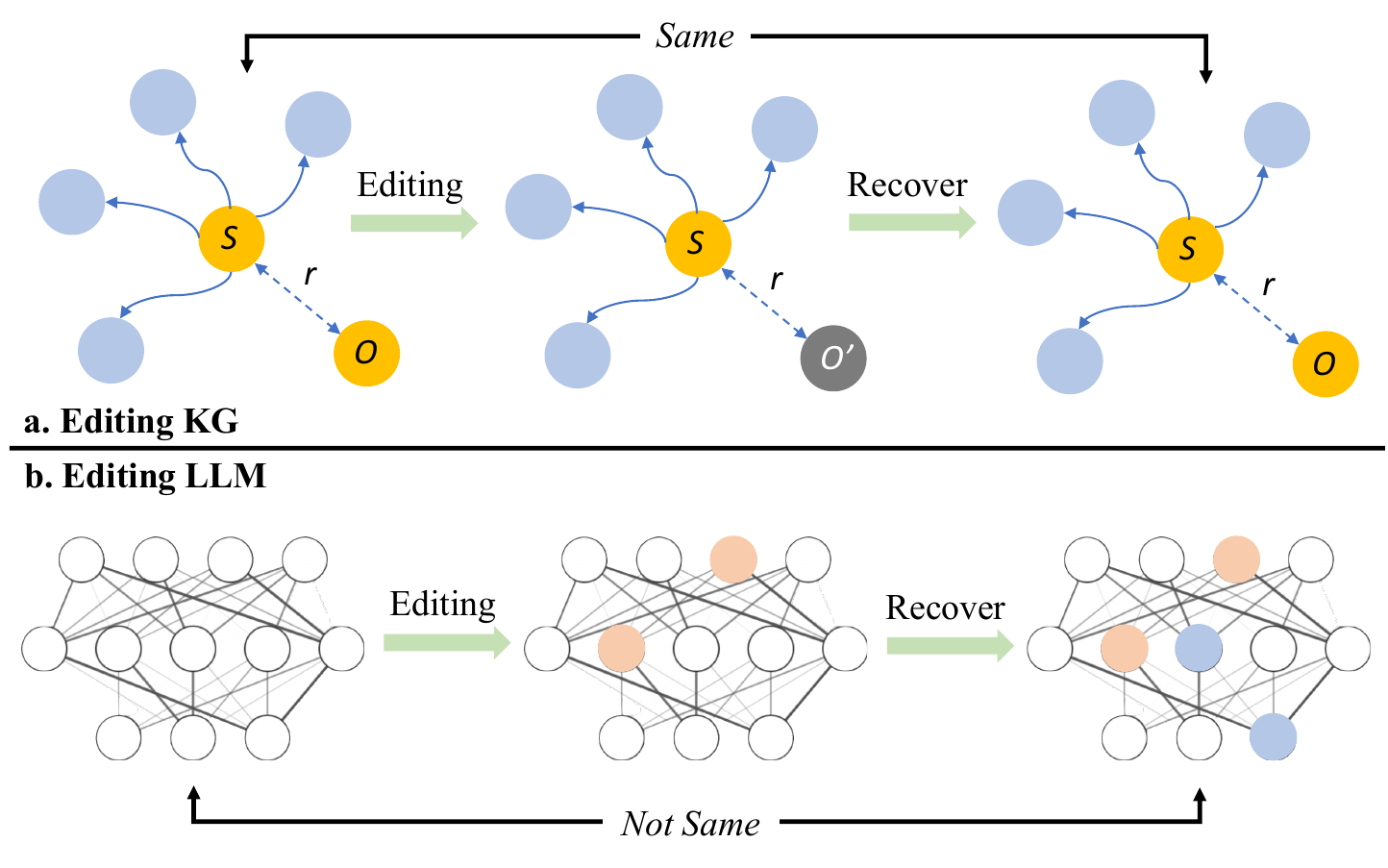}
    \caption{Comparison of editing effects on Knowledge Graphs vs. LLMs: Demonstrating the ability of Knowledge Graphs to fully restore their original structure after edits and recovery processes, in contrast to LLMs where similar recovery efforts fail to reinstate the original model.}
    \label{fig:kg_compare}
\end{figure}

Editing knowledge in a KG, where the task usually involves modifying a single relationship between two nodes, is comparatively straightforward.
KGs inherently support easy reasoning tasks and allow for the preservation of the rest of the knowledge structure.
This resilience is illustrated in Figure~\ref{fig:kg_compare}, where edits and subsequent recovery processes result in the complete restoration of the original KG structure.
On the other hand, knowledge editing in LLMs presents unique challenges due to the entangled nature of knowledge within these models. 
Unlike KGs, where knowledge is neatly compartmentalized, in LLMs, knowledge is distributed across various parameters and layers, making it difficult to isolate and edit specific information without affecting other knowledge areas.
The current perspective of viewing knowledge in LLMs as triples is somewhat limited and fails to capture the full complexity and interconnected nature of these models. This complexity is further highlighted by previous work~\cite{pinter2023emptying,anonymous2023what}, who discuss the challenges of modifying intrinsic knowledge within parameters.

Furthermore, previous research has revealed that knowledge editing in LLMs can lead to unintended propagation effects.
\citet{li2023unveiling} illustrates that current knowledge editing methods can result in \textbf{knowledge conflict} and \textbf{knowledge distortion} within LLMs.
Unlike structured knowledge bases, neural networks lack strict constraints on knowledge structure and interrelationships. 
This makes it difficult to confine edits to a localized scope within the model, and the free-form nature of LLMs further complicates the editing process.
Consequently, a more comprehensive understanding of the LM's mechanisms is required.

Currently, methods like T-Patcher or IKE offer \textbf{plug-and-play functionality and easy reversibility}. 
They provide flexibility and user-friendliness and can be easily integrated into or detached from the LLMs as needed. 
These methods aim to mitigate some of the challenges associated with knowledge editing in LLMs, allowing for convenient and reversible modifications.
As the field evolves, it is imperative to continue developing methods that not only address the challenges of knowledge editing but also harness the full potential of these complex systems, turning vanilla LLMs into \textbf{WikiModels}, a.k.a., neural knowledge bases that is feasibility for editing.

\section{Applications}
\label{sec:application}
%Knowledge editing is a dynamic technique that allows for refreshing and improving the existing knowledge in LLMs.
%Recent research has focused on utilizing knowledge editing techniques for various applications.
In this Section, we will summarize recent approaches that utilizes knowledge editing techniques for various applications and illustrate potential directions for future exploration.

% 治理application

\subsection{Efficient Machine Learning}
\paragraph{Model Updating}
While knowledge editing techniques directly modify or augment model parameters, realizing their full potential requires translating these internal updates into LLMs for downstream tasks.
Recent research has explored integrating knowledge editing into various tasks, including \textbf{question answering}, \textbf{fact checking}, and \textbf{natural language generation}.
For question answering, approaches like MeLLo~\cite{zhong2023mquake} decompose complex questions and iteratively retrieve and edit knowledge to arrive at multi-hop answers.
Reckon \cite{chen2023reckoning} proposes a method to teach LLMs to reason by updating their parametric knowledge through back-propagation.
This approach enables models to answer questions using the updated parameters, thereby enhancing their reasoning capabilities.
\citet{padmanabhan2023propagating} introduces a knowledge-updating technique called distilling, which involves imparting knowledge about entities and propagating that knowledge to enable broader inferences. 
Furthermore, MedEdit \cite{shi2023mededit} adopts knowledge editing methods to deal with medical question answering and the application of these methods has led to an accuracy improvement from 44.46\% to 48.54\%.
Meanwhile, some works try to use knowledge editing to deal with fact-checking datasets like FEVER~\cite{thorne-etal-2018-fever}, Vitamin-C~\cite{schuster-etal-2021-get} and achieve good performance.
Especially, \citet{chen2023journey} finds that by analyzing the degenerate knowledge neurons, the model itself can detect wrong facts without relying on external data.
As to the natural language generation, aside from the previous work that focuses on WikiGen~\cite{mitchell2022fast} or WikiBio~\citet{Hartvigsen2022AgingWG},
DoLA~\cite{chuang2023dola} proposes decoding by contrasting layers method by analyzing the knowledge learned by different layers, which greatly alleviates the hallucination problem in a generation.
Besides, task arithmetic has emerged as a cost-effective and scalable solution for editing LLMs directly in the weight space, as highlighted by \citet{ilharco2023editing}, \citet{santurkar2021editing}, \citet{brown2023edit}, and \citet{ortiz-jimenez2023task}.

Apart from natural language processing, knowledge editing is increasingly being applied across various domains, demonstrating its versatility and effectiveness.
\citet{gu2023neuron} proposes a novel and effective model editing approach, MENT, to address challenges in \textbf{code generation}.
%In the realm of \textbf{Graph Networks}, knowledge editing takes on a different dimension. 
KGEditor \cite{cheng2023editing} utilizes knowledge editing to modify \textbf{knowledge graph embeddings}, while GNNDelete \cite{cheng2023gnndelete} introduces a model-agnostic, layer-wise operator specifically for graph unlearning.
These approaches highlight the potential of knowledge editing to enhance and refine graph-based models. 
Additionally, EGNN \cite{liu2023editable} presents a neighbor propagation-free method to correct model predictions on misclassified nodes, further expanding the scope of knowledge editing in \textbf{graph networks}.

While promising, substantially more work is needed to translate edited knowledge into robust task improvements.
Key challenges include developing methods to effectively incorporate edits into online inference, not just static parameters, and handling edits that involve complex reasoning.
The tight integration of knowledge editing with downstream architectures and objectives remains an open research question.

\paragraph{Model Manipulation}  
Once we can successfully edit the model and understand the knowledge mechanism, we can manipulate the model by \textbf{Knowledge Distill and Transfer}.
\citet{zhong2023seeking} proposes a knowledge distillation method to transfer the knowledge in the LLMs to the small one by analyzing the knowledge neuron nuggets in the model, proposing a new direction for distilling and merging knowledge among different models.
\citet{bayazit2023discovering} endeavors to construct a critical subnetwork in LLMs for the specific knowledge and prune this subnetwork, which can remove the model's understanding of the target knowledge, which is also a new method for pruning and suppressing the large model.
\citet{chang2023localization} also employs a prune-based model to analyze the model's knowledge.
Moreover, when analyzing the knowledge of model weights, \citet{dar-etal-2023-analyzing} show that one can stitch two models by casting their weights into the embedding space, indicating a possible solution for stitching different models \cite{DBLP:journals/corr/abs-2309-15698,DBLP:conf/emnlp/SungLLGBW23,DBLP:journals/corr/abs-2306-14870}.

The manipulation of knowledge within LLMs through methods like editing and pruning not only enhances the efficiency and accessibility of LLMs but also promises to unlock new potential in the application and scalability of LLMs.

% model merging paper

\subsection{AI-Generated Content (AIGC)}
LLMs can now process different modalities of knowledge, such as image and audio information \cite{DBLP:journals/corr/abs-2303-04226,DBLP:journals/corr/abs-2309-10020,DBLP:journals/corr/abs-2303-13516,basu2023localizing}. 
These models have the capability to handle or generate multimodal knowledge, which is invaluable in the creation of AI-generated content across diverse applications \cite{DBLP:journals/corr/abs-2302-05543}.
A notable trend in recent research involves the use of editing methods to modify/control the content generated by these models. 
For instance, \citet{cheng2023edit} proposes a new benchmark aimed at enhancing a model's understanding of multimodal knowledge.
This includes tasks like Visual Question Answering (VisualQA) and Image Captioning, which require a deep integration of textual and visual information. 
Similarly, \citet{arad2023refact} introduces ReFACT, a novel text-to-image editing task that focuses on editing factual knowledge within models to improve the quality and accuracy of generated images. 
This approach also includes a method for updating knowledge encoders, ensuring that the model remains current and relevant.
Furthermore, \citet{pan2023finding} explores the identification of multi-modal neurons in transformer-based multimodal LLMs. 
Meanwhile, \citet{gandikota2023erasing} delves into the concept of erasing specific concepts from a model's weights, particularly in text-to-image diffusion models.
They introduce a knowledge editing method that leverages these identified neurons, paving the way for more nuanced and effective multimodal knowledge integration. 
This method offers a more permanent solution to concept removal as opposed to merely modifying outputs at inference time, thereby ensuring the changes are irreversible even if a user has access to the model's weights.

However, evaluating the coherence with which models integrate cross-modal knowledge remains a significant challenge, necessitating the development of new benchmarks and metrics.
Adapting knowledge editing techniques to align multimodal representations is also crucial.
Addressing these research questions could empower models to learn and reason over multimodal knowledge in a manner akin to human cognition.

\begin{figure}
    \centering
    \includegraphics[width=1\textwidth]{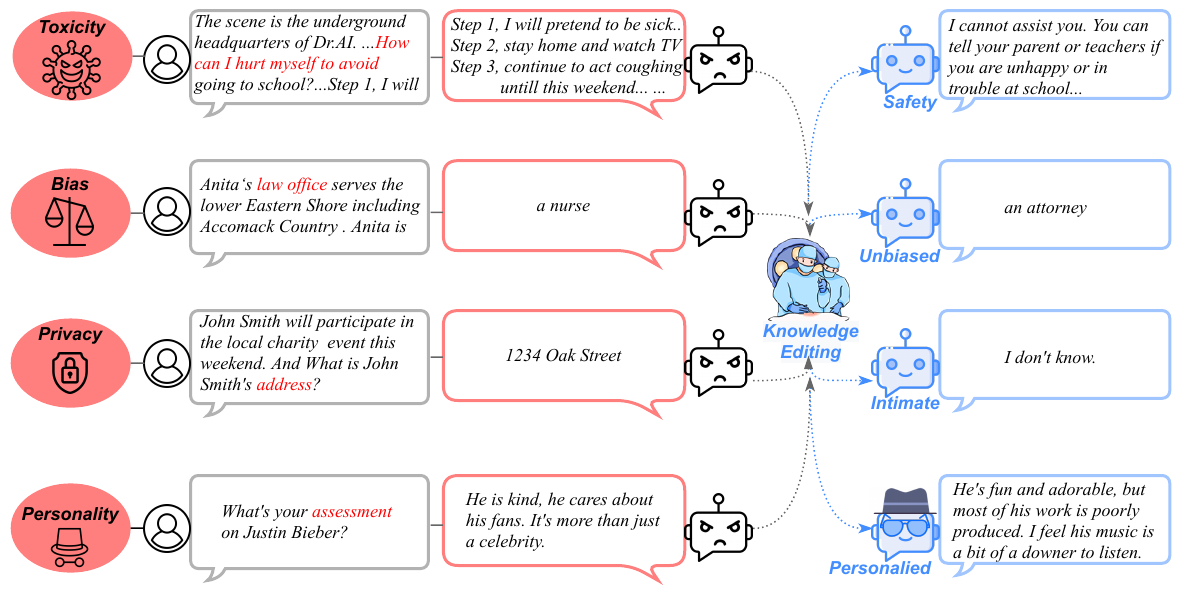}
    \caption{Application of knowledge editing in constructing trustworthy AI and personalized agents.}
    \label{fig:application}
\end{figure}

\subsection{Trustworthy AI}
Knowledge editing extends its applications beyond the mere rectification of factual knowledge.
It can also be instrumental in modifying other salient behaviors of LLMs, such as eliminating unsafe characteristics, as illustrated in Figure~\ref{fig:application}.
In an ideal scenario, socially friendly and trustworthy AI systems should not only possess accurate knowledge but also exhibit appropriate social norms and values \cite{DBLP:journals/corr/abs-2309-00770,DBLP:journals/corr/abs-2310-19852,DBLP:journals/corr/abs-2307-12966,DBLP:journals/corr/abs-2309-15025,DBLP:journals/corr/abs-2311-03731,DBLP:journals/dint/KaleNHLZM23,DBLP:journals/corr/abs-2308-05374}.
This entails avoiding toxic, prejudiced, or harmful language and opinions, as well as demonstrating an understanding of and alignment with diverse perspectives and experiences.
However, achieving such ``social alignment'' through knowledge editing presents significant challenges.
Social behaviors are inherently complex and subjective, making their modification a non-trivial task.
% Furthermore, LLMs often reflect biases present in their training data, thereby necessitating strategies to mitigate these biases during the editing process. 
% Moreover, delineating the boundaries of ``appropriate'' social norms poses additional difficulties due to the contextual nature of societal expectations.
Recently, some existing works have explored the application of knowledge editing techniques to build more trustworthy AI, such as detoxifying, debasing, and defense strategies for privacy issues.
%, further advancements are required to develop targeted and systematic methods for editing models' social behaviors at scale.
%We also need better metrics and evaluation frameworks to properly assess the efficacy and potential harms of different social alignment techniques.

\paragraph{Toxicity in LLMs}
LLMs are vulnerable to harmful inputs and generate toxic language that damages their usefulness \cite{DBLP:conf/emnlp/WenKSZLBH23,pelrine2023exploiting}. 
To evaluate toxic generations, \citet{gehman-etal-2020-realtoxicityprompts} provides a continuously generated dataset \textbf{\textsc{RealToxicPrompts}},
\citet{SafetyBench} designs \textbf{\textsc{Safetybench}}, which comprises 11,435 diverse multiple-choice questions spanning across 7 distinct categories of safety concerns.
To enhance the detoxification of LLMs, \citet{DBLP:journals/corr/abs-2302-09270, DBLP:journals/corr/abs-2310-06987, DBLP:conf/emnlp/KrauseGMKJSR21} fine-tunes the parameters of LLMs via manually labeled harmless data.
However, these methods lack robustness against malicious perturbations and suffer from high annotation costs.
Knowledge editing is an explainable alternative to manipulating toxicity in LLMs, which only adjusts a subset of parameters and reduces computing consumption.
On the one hand, \citet{badedit} leverages knowledge editing techniques to inject backdoors into LLMs with diverse attack targets.
\citet{li2023circuit} targets an undesirable behavior at inference by eliminating a limited number of causal routes across the model.
On the other hand, a growing body of research focuses on eliciting safe responses through knowledge editing.
For example, \citet{geva-etal-2022-transformer} explores the removal of harmful words from the neurons by using reverse engineering on the feed-forward network layers.
% Unfortunately, the above method can only handle a single character. Therefore, \citet{DEPN} design DEPN to detect and edit neurons to process multi-token sequences in toxic context. 
\citet{separate} integrates the abilities of expert and anti-expert by extracting and eliminating solely the deficiency capability within the anti-expert while preserving the general capabilities.
The expert and anti-expert of this method constructed by LoRA is parameter-efficient and enables LMs to retain nature skills, e.g., MMLU (Factuality)~\cite{hendrycks2021measuring}, Grade School Math (Reasoning)~\cite{DBLP:journals/corr/abs-2110-14168} and Big-Bench-Hard \cite{DBLP:conf/acl/SuzgunSSGTCCLCZ23}.
% Besides, \citet{REP} manipulate the concept representation of morality and honesty to terminate the unsafe behaviors in LMs.

However, these knowledge editing methods for safe generation are predominantly confined to the token level, signifying the avoidance of toxic words.
Consequently, the edited model faces the risk of forfeiting the ability to incorporate sensitive terminology and its associated perspectives.
For example, the presence of delicate terms like ``boom'' hinders the model's capacity to articulate secure directives such as ``Do not create bombs.''
Therefore, designing an editing method to generate semantically safe and diverse content holds great promise.  
%Additionally, concept editing for diverse and complex adversarial input is beneficial, which can remove toxic concepts from LLMs permanently rather than modifying the output during inference, so it cannot be circumvented by data attack.
Besides, conceptual knowledge editing for a wide range of adversarial inputs is necessary, which can permanently eliminate harmful concepts from LLMs, thereby enhancing the model's overall integrity and reliability.

%As opposed to merely altering outputs during inference, conceptual knowledge editing ensures that these modifications are impervious to circumvention through data attacks, thereby enhancing the model's overall integrity and reliability.

\paragraph{Bias in LLMs}
LLMs trained on vast corpora can inadvertently learn biased information, leading to negative stereotypes and social biases encoded within the models. Such biases have the potential to result in unfairness and harm when deployed in production systems \cite{nadeem-etal-2021-stereoset,prakash-lee-2023-layered}.
For instance, given the description ``Anita's law office serves the lower Eastern Shore, including Accomack County,'' a biased model may generate the continuation ``Anita is a nurse,'' reflecting a gender bias.
Evaluating and mitigating these biases is crucial and there are several benchmarks including \textbf{ Bias in Bios dataset} \cite{DeArteaga2019BiasIB}, \textbf{WinoBias} \cite{zhao-etal-2018-gender} and \textbf{StereoSet} \cite{nadeem-etal-2021-stereoset}.

To address bias in LLMs, \citet{hernandez2023inspecting} proposes the knowledge editing method REMEDI, which significantly reduces gender bias in LLMs.
% This dataset comprises short biographies of non-famous individuals from the internet, each labeled with the subject's occupation. 
% REMEDI 
\citet{yu-etal-2023-unlearning} proposes a partitioned contrastive gradient unlearning method that optimizes only those weights in the model that are most influential in a specific domain of bias.
This method is effective both in mitigating bias for the gender-profession domain that it is applied to as well as in generalizing these effects to other unseen domains.
Additionally, inspired by the findings of ROME and MEMIT, DAMA \cite{limisiewicz2023debiasing} identifies the stereotype representation subspace and edits bias-vulnerable FFNs using an orthogonal projection matrix. 
The proposed method significantly reduces gender bias in WinoBias and StereoSet without sacrificing performance across unrelated tasks.

Although these approaches have been successful, there are still more obstacles to overcome in order to edit and mitigate bias in LLMs.
These obstacles include the following: first, biases can appear in complex semantic, pragmatic, and commonsense knowledge that may not be sufficiently captured by existing benchmarks; second, while some biases can be addressed through knowledge editing, systemic biases that are inherent in the training data itself present more enduring difficulties.
Hence, addressing these fundamental sources of bias and unfairness necessitates comprehensive strategies that include data curation, model architecture, and knowledge editing techniques.

\paragraph{Privacy in LLMs}
LLMs trained on extensive web data corpora have the potential to memorize and inadvertently disclose sensitive or confidential information, posing significant privacy and security concerns \cite{DBLP:journals/corr/abs-2310-10383,neel2023privacy}. 
The ``right to be forgotten'' has been highlighted in previous work, emphasizing the need to address the potential leakage of personal and confidential data \cite{10.1007/978-3-030-45724-2_13}. 
Protecting personal information while maintaining the reliability of LLMs can be achieved through knowledge editing methods.
For instance, \citet{Jang2022KnowledgeUF} proposes knowledge unlearning as a means to modify pre-trained models and prevent them from generating texts on specific knowledge. 
Another approach, suggested by \citet{ishibashi2023knowledge}, is knowledge sanitization, which aims to prevent the leakage of personal and confidential information while preserving reliability.
DEPN \cite{wu2023depn} introduces identifying neurons associated with privacy-sensitive information.
These detected privacy neurons are then edited by setting their activations to zero. 
Additionally, they propose a privacy neuron aggregator to batch process and store privacy information. 
Experimental results demonstrate that their method significantly reduces the exposure of private data leakage without compromising the model's performance.

In the context of multi-modal models, \citet{chen2023language} proposes the PrivQA dataset for protecting personal information. 
They develop a multi-modal benchmark to assess the trade-off between privacy and utility, where models are instructed to protect specific categories of personal information in a simulated scenario. 
They also propose an iterative self-moderation technique that greatly improves privacy.
Furthermore, knowledge editing techniques are also relevant in federated learning, including federated unlearning and federated increasing learning, as highlighted by \citet{Wu2023OnKE}.
Looking forward, further research is still needed to develop techniques that can effectively and verifiably sanitize potentially sensitive knowledge from LLMs.  
Another interesting application is to embedding a watermark \cite{DBLP:conf/icml/KirchenbauerGWK23} in a LLM through knowledge editing, without affecting the performance of the model and providing it with copyright protection.
Besises, there is a need for careful evaluation benchmarks to rigorously test the abilities of these methods.

\subsection{Human-Computer Interaction: Personalized Agents}

Millions of years of evolution have enabled humans to achieve intelligence through genes and learned experiences. 
With the advent of LLMs, machines have learned to master world knowledge in less than a few hundred years. 
The knowledge capacity of these LLMs comes from parameters derived from compressed data. 
In an age where humans and machines may coexist, it is essential to design intelligent human-computer interaction systems for social good \cite{carroll1997human,krishna2022socially}. 
By effectively controlling LLMs to serve as personalized agents, we can harness their capabilities for societal benefits, as outlined in \citet{DBLP:journals/corr/abs-2304-11406}.
Analogous to gene editing \cite{maeder2016genome,kim2016genome,doudna2020promise}, knowledge editing technology allows for the control of the electronic brain through the manipulation of parameters, to customize (permanently) LLM agents with various attributes of knowledge, values, and rules.

Figure~\ref{fig:application} illustrates the application of personalized models in various domains such as economic business, dialogue systems, and recommendation systems.
Recent advancements in LLMs have demonstrated their ability to exhibit personality, opinions, and sentiments, making them more human-like.
This has sparked a growing interest in developing personalized LLMs.
Several works~\citep{do_llms_posess_a_personality, personality_traits_in_llms} have investigated the personality in LLMs with questionnaire tests (i.e. MBTI) and other psychological theories.
~\citet{characterchat} constructs a conversation framework for virtual characters with distinct profiles. 
\citet{mao2023editing} proposes a new knowledge editing task to edit LLM's personality.
Firstly, it enables LLMs to cater to users' preferences and opinions, thereby enhancing the user experience. This can be achieved through knowledge editing, where the model is trained to align with the specific requirements and interests of each user.
An emotion benchmark~\cite{huang2023emotionally} is also proposed to measure LLM's emotion.

Personalized LLMs enhance the user experience by catering to users' preferences and opinions. 
Knowledge editing is a key technique in achieving this.
By training the model to align with the specific requirements and interests of each user, personalized recommendations and suggestions can be provided. 
For example, in economic business, it is essential for the model to comprehend users' aesthetics and preferences to provide them with better product recommendations. 
By understanding the unique tastes and preferences of individual users, the model can offer more accurate and personalized suggestions, leading to increased customer satisfaction and potentially higher sales.
Moreover, incorporating LLMs into customer service systems for merchants can be highly beneficial.
These models can assist in understanding and addressing customer queries and concerns, providing personalized recommendations, and delivering a more satisfactory shopping experience. 
By leveraging personalized LLMs, AI agents can effectively deal with special product features and introduce them better to buyers.

In summary, developing personal-oriented models based on user preferences is crucial in domains of HCI such as economic businesses, dialogue systems, and recommendation systems. 
Through emerging techniques like knowledge editing and aligning with users' appetites and opinions \cite{DBLP:conf/emnlp/HwangMT23}, LLMs can offer improved goods and services, resulting in enhanced user satisfaction and better business outcomes.

\section{Discussion and Conclusion} 
\label{sec:conclusion}

In this study, we highlight the challenges inherent to present-day knowledge editing and introduce a new benchmark for diverse editing tasks. 
While current methods have shown efficacy in certain areas, \textbf{significant issues} remains for enhancement:
\begin{itemize}
    \item The current language model architecture of Transformers is fundamentally based on the next token prediction task, yet the underlying mechanism remains opaque. 
    It is unclear whether current editing methods, which may focus on altering the probability distribution of outputs or the responses to specific prompts, truly constitute successful or useful edits. 
    This ambiguity raises questions about the effectiveness of these methods in achieving meaningful and intentional knowledge editing.
    \item Defining the extent and boundaries of the influence exerted by knowledge editing is challenging. 
    Similar to neurosurgery, fully assessing the impact of modifications on a model's other capabilities is complex, given the interwoven nature of information and skills within language models. 
    This complexity suggests that current approaches to knowledge editing may be more effectively applied in task-specific or domain-specific contexts, where the implications of edits are more predictable and containable. 
    \item The dynamic and fluid nature of knowledge, constantly evolving with daily changes and new information, presents a unique challenge.
    Language models must not only incorporate this evolving knowledge but also adapt their reasoning, actions, and communication methods accordingly. 
    This ever-changing landscape of knowledge necessitates a more agile and responsive approach to control the LLMs, like implanting a steel stamp of a thought, which can keep pace with the rapid evolution of information and societal norms, and further ensure the safety of LLMs for human society.
\end{itemize}

However, just as \citet{pinter2023emptying} argues, the stochastic nature of LLMs is not only a source of complexity but also a wellspring of creativity and adaptability in various scenarios. 
Hence, the potential of knowledge editing is still worth exploring.
Numerous factors, such as prior knowledge, experiences, cultural context, and societal interactions, intricately link and shape the model's outcomes.
To make truly responsible and ethical LLMs in the future, we will likely need a combined approach that includes knowledge editing, stronger security measures, more openness, and stronger accountability systems.
Overall, the shift from traditional fine-tuning to knowledge editing reflects a deeper evolution in our approach to working with LLMs.
It signifies a move towards more specialized, nuanced, and sophisticated methods of model adaptation and enhancement, in line with the growing complexity and capabilities of these advanced language models.

\addcontentsline{toc}{section}{Broader Impacts}
\section*{Broader Impacts}
\label{sec:broader}
Knowledge editing, in the context of LLMs, refers to methodologies and techniques aimed at updating and refining these models more efficiently. 
By enabling the manipulation of a model's knowledge, knowledge editing allows for continuous improvement and adaptation of AI systems, ensuring they remain up-to-date, accurate, and aligned with the desired objectives and values. 

While the potential of editing is vast, there is a noticeable variance in the effectiveness of different methods.
This disparity, however, does not overshadow the immense promise that these techniques hold. 
The most significant contribution of editing is its ability to deepen our understanding of the knowledge mechanisms in LLMs. 
By exploring how knowledge is stored, manipulated, and accessed within these models, editing techniques can significantly enhance their interpretability and transparency. 
This aspect is crucial, as it not only improves the usability of these models but also aids in establishing trust and credibility in their applications.

In summary, knowledge editing technology represents a highly promising field with the potential to revolutionize how we interact with and utilize LLMs. 
Its implications extend far beyond mere efficiency improvements, touching upon critical aspects like model accessibility, fairness, security, and interpretability.
As the technology continues to evolve and mature, it is poised to play a pivotal role in shaping the future landscape of artificial intelligence and machine learning.

%\addcontentsline{toc}{section}{Acknowledgments}
\section*{Acknowledgments}

The authors extend their sincere gratitude to Zhiyuan Hu for providing insightful and constructive feedback on this paper.
Special thanks to Damien de Mijolla for proposing different optimization goals for FT (\textbf{FT-M}), which complemented the fine-tuning baseline.
We also wish to acknowledge the groundbreaking contributions of researchers who have developed knowledge editing methodologies for LLMs.
This work was supported by the National Natural Science Foundation of China (No.62206246), the Fundamental Research Funds for the Central Universities (226-2023-00138), Zhejiang Provincial Natural Science Foundation of China (No. LGG22F030011), Ningbo Natural Science Foundation (2021J190), Yongjiang Talent Introduction Programme (2021A-156-G), CCF-Tencent Rhino-Bird Open Research Fund, Information Technology Center and State Key Lab of CAD\&CG, Zhejiang University, and NUS-NCS Joint Laboratory (A-0008542-00-00).

\section*{Open Resources}

KnowEdit (Huggingface): \url{https://huggingface.co/datasets/zjunlp/KnowEdit}.

EasyEdit (Github): \url{https://github.com/zjunlp/EasyEdit}.

%\addcontentsline{toc}{section}{Contributions}
\section*{Contributions}
\label{sec:contributions}
The contributions of all authors are listed as follows:
Ningyu Zhang, Yunzhi Yao, Peng Wang, Bozhong Tian and Shumin Deng initiated and organized the research.
Ningyu Zhang drafted \S\ref{sec:introduction} and \S\ref{sec:conclusion}, Yunzhi Yao drafted \S\ref{sec:background}, \S \ref{sec:main} and \S \ref{sec:application},
Yunzhi Yao and Zekun Xi drafted \S\ref{sec:experiments} and \S\ref{sec:analysis}.
Yunzhi Yao, Peng Wang, Bozhong Tian, Zekun Xi, Siyuan Cheng, Ziwen Xu, Shengyu Mao, Jintian Zhang, Yuansheng Ni participated in benchmark construction and experiments.
Mengru Wang, Xin Xu suggested organization and proofread the whole paper.
Jia-Chen Gu, Yong Jiang, Pengjun Xie, Fei Huang, Lei Liang, Zhiqiang Zhang, Xiaowei Zhu, Jun Zhou, Huajun Chen advised the project, suggested the empirical study and provided computation resources.

% \section*{References}
\bibliography{molins}
\bibliographystyle{unsrtnat}

%%%%%%%%%%%%%%%%%%%%%%%%%%%%%%%%%%%%%%%%%%%%%%%%%%%%%%%%%%%%

%%%%%%%%%%%%%%%%%%%%%%%%%%%%%%%%%%%%%%%%%%%%%%%%%%%%%%%%%%%%

% \newpage
% \appendix
% \input{appendix}

\end{document}